\documentclass{article}

\usepackage{microtype}
\usepackage{graphicx}
\usepackage{subfigure}
\usepackage{booktabs} 
\usepackage{bm}
\usepackage{flushend}

\usepackage{enumitem}
\usepackage{amsmath}
\usepackage{multirow}
\usepackage{array}
\usepackage{makecell}
\usepackage{amsfonts}       
\usepackage{nicefrac}       
\usepackage{appendix}
\newcolumntype{P}[1]{>{\centering\arraybackslash}p{#1}}
\usepackage{hyperref}
\usepackage{cleveref}
\usepackage{listings}
\newcommand{\edited}[1]{{\color{blue} #1}}

\crefformat{section}{\S#2#1#3} 
\crefformat{subsection}{\S#2#1#3}
\crefformat{subsubsection}{\S#2#1#3}

\usepackage[accepted]{mlsys2023}

\usepackage[firstpage]{draftwatermark}
\SetWatermarkText{
 \hspace*{3.5in}
 \raisebox{10.12in}{
  \includegraphics[height=0.9in]{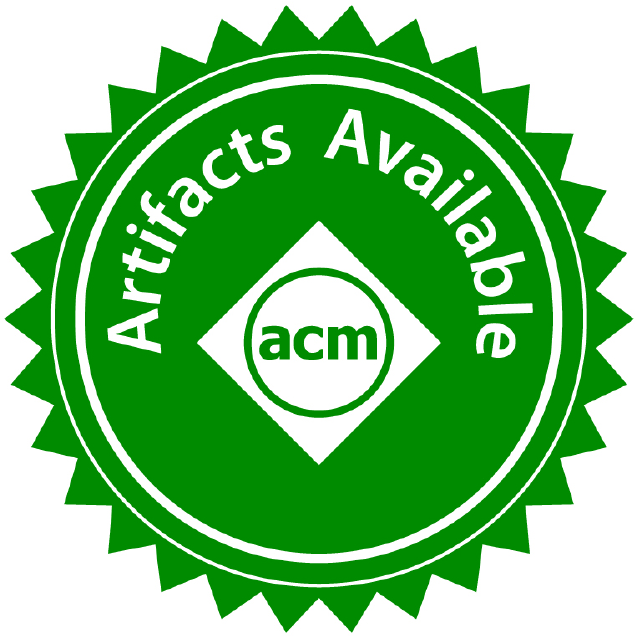}
  \includegraphics[height=0.9in]{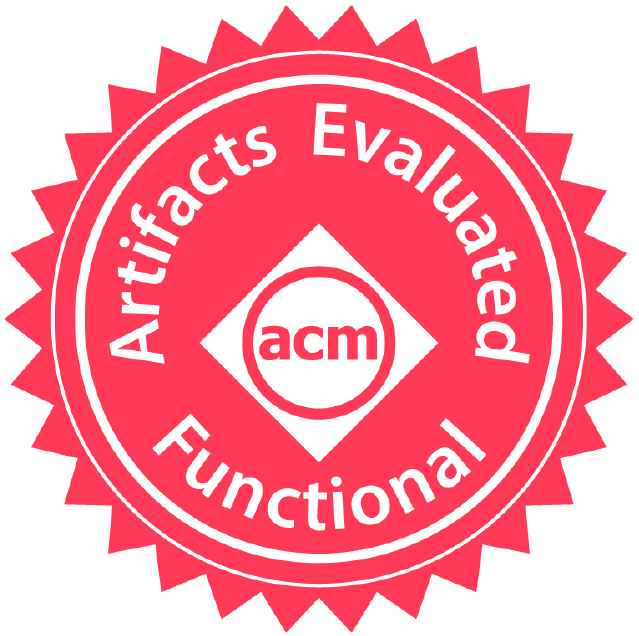}
 }
}
\SetWatermarkAngle{0}

\begin{document}

\twocolumn[
\mlsystitle{GiPH: Generalizable Placement Learning for Adaptive Heterogeneous Computing}



\mlsyssetsymbol{equal}{*}

\begin{mlsysauthorlist}
\mlsysauthor{Yi Hu}{cmu}
\mlsysauthor{Chaoran Zhang}{cmu}
\mlsysauthor{Edward Andert}{asu}
\mlsysauthor{Harshul Singh}{cmu}
\mlsysauthor{Aviral Shrivastava}{asu}
\mlsysauthor{James Laudon}{goo}
\mlsysauthor{Yanqi Zhou}{goo}
\mlsysauthor{Bob Iannucci}{goo}
\mlsysauthor{Carlee Joe-Wong}{cmu}
\end{mlsysauthorlist}

\mlsysaffiliation{cmu}{Department of Electrical and Computer Engineering, Carnegie Mellon University, Pittsburgh, Pennsylvania, USA}
\mlsysaffiliation{goo}{Google Brain, Mountain View, California, USA}
\mlsysaffiliation{asu}{School of Computing and Augmented Intelligence, 	Arizona State University, Tempe, Arizona, USA}

\mlsyscorrespondingauthor{Yi Hu}{yihu@andrew.cmu.edu}


\mlsyskeywords{Reinforcement Learning, heterogeneous computing, placement}

\vskip 0.3in

\begin{abstract}
Careful placement of a distributed computational application within a target device cluster is critical for achieving low application completion time.
The problem is challenging due to its NP-hardness and combinatorial nature. In recent years,  learning-based approaches have been proposed to learn a placement policy that can be applied to unseen  applications, motivated by the problem of placing a neural network across cloud servers. These approaches, however, generally assume the device cluster is fixed, which is not the case in mobile or edge computing settings, where heterogeneous devices move in and out of range for a particular application. {To address the challenge of scaling to different-sized device clusters and adapting to the addition of new devices, }we propose a new learning approach called \textbf{GiPH}, which learns policies that generalize to dynamic device clusters via 1) a novel graph representation \textbf{gpNet} that efficiently encodes the information needed for choosing a good placement, and 2) a scalable graph neural network (GNN) that learns a summary of the gpNet information. GiPH turns the placement problem into that of finding a sequence of placement improvements, learning a policy for selecting this sequence that scales to problems of arbitrary size. We evaluate GiPH  with a wide range of task graphs and device clusters and show that our learned policy rapidly finds good placements for new problem instances. GiPH finds placements that achieve up to $30.5\%$ better makespan, searching up to $3\times$ faster than other search-based placement policies. 

\end{abstract}
]



\printAffiliationsAndNotice{}  

\section{Introduction}


 When running a compute application across a network of computing devices, careful choice of which parts of the application to run on which device can significantly affect application performance. This is particularly true when devices are heterogeneous: e.g., compute-intensive tasks should be run on devices with more computation resources, unless those devices have insufficient communication resources to transmit the task results. {Moreover, for applications involving mobile entities (e.g., autonomous vehicles, mobile users) devices can be volatile: they may unexpectedly enter and/or exit the system. {Their capabilities may also vary, e.g., due to competing processes or battery drainage.}} Finding an adaptive placement solution is therefore challenging. \citet{salaht2020overview}, for example, survey prior works on service placement in edge computing settings, which feature a \emph{heterogeneous} and \emph{dynamic} mix of edge devices, edge servers, and cloud servers. 
 
 We consider applications {in heterogeneous computing that must adapt 
{to device network changes}. One example is cooperative sensor fusion for intelligent traffic systems, where vehicles combine sensor data from other cars and traffic cameras for localization and autonomous driving. In general, the applications} can be represented as directed acyclic graphs (DAGs), in which each node represents a discrete computation task or sensor input, and edges represent data links between tasks that determine the sequence of computations. Input data enters the DAG and flows through the tasks, e.g., {camera images moving through a trained convolutional neural network (CNN) for object detection.} 
 We focus on the common objective of minimizing application completion time, e.g., to receive CNN inference results as fast as possible. {Our proposed framework, however, can generalize to other objectives, e.g., if one wishes to balance energy cost with completion time.}

\paragraph{Challenges:} Due to the NP-hardness and combinatorial nature of the {placement} problem~\citep{heft}, heuristic methods that rely on simple strategies and hand-crafted features have been proposed. However, they are often sub-optimal. 
Many heuristics also assume an overly simplified performance model (e.g., for task computation and communication times) to enable a closed-form formulation. This paper, instead, follows another recent line of work that automatically learns highly efficient placement policies with reinforcement learning (RL), which learns directly from simulated or real runtime environments. 

RL also has the potential to generalize across different problem instances, which can reduce the need for re-training when new applications need to be placed on new clusters of devices. 
In practice, \emph{a wide range of applications may run on constantly changing networks of devices}, e.g., training a federated learning model~\cite{tu2020network}, rendering augmented reality holograms~\citep{wang2018service}, or analyzing camera video feeds~\citep{hoque2021r}, all of which may be (partially) run on mobile devices like phones or vehicle-mounted cameras whose availability changes over time as they move~\citep{salaht2020overview}. It is thus imperative to design RL representations and learning algorithms that \emph{can adapt placements to new device clusters}. Existing  approaches cannot do so as they either assume all device characteristics are fixed 
or a fixed number of devices. As a result, whenever the device network changes, the learned policies will perform poorly without significant re-training. In heterogeneous computing environments, placement feasibility constraints may also exist due to hardware functionality, {which in general can be challenging to handle in RL}.

In this work, we propose \textbf{GiPH}, an RL-based approach to learning efficient and fully Generalizable Placement with the ability to adapt to dynamic Heterogeneous networks. To the best of the authors' knowledge, GiPH is the very first {RL approach} to learn a placement policy that not only generalizes to new task graphs {that are not in the training set}, but also explicitly adapts to changing device networks { with the ability to \textit{relocate} the tasks}. While we consider minimization of completion time, GiPH generalizes to other performance objectives. After reviewing related work in Section~\ref{sec:related}, we outline the \textbf{key contributions} and findings of this work as follows:
\begin{itemize}[leftmargin=*]
    \item We formulate application placement as a \textit{search problem} where the placement is found through applying a sequence of iterative {task relocations that improve upon a given placement (Section~\ref{sec:problem}).}
 \item We propose \textbf{GiPH}, an RL-based framework to learn \textit{generalizable} placement policies. The learned policy efficiently searches for good placement of a task graph on a heterogeneous device cluster {using a novel and \textit{universal} graph representation called \textbf{gpNet} {that explicitly encodes task-level and device-level information}} (Section~\ref{sec:giph}). 

    
\item  We \emph{evaluate GiPH} in terms of the placement quality and generalizability.  GiPH finds placements that achieve up to 30.5\% lower completion times, with higher search efficiency than prior work.  It is comparable to HEFT, a state-of-the-art but slow~\cite{gcnscheduler} heuristic specific to completion times (Section~\ref{sec:eval}).

    \item {We present a case study on applying GiPH to a complex real-world application using cooperative sensor fusion and demonstrate the practical effectiveness of GiPH.}
\end{itemize}





\section{Related Work}\label{sec:related}
Application placement is relevant to a variety of domains. GiPH is inspired by the line of work on efficient device placement for distributed neural network training with RL~\citep{dp_rl, Mirhoseini2018AHM_hdp, gdp_google, Paliwal2019REGALTL, addanki2019placeto}, where a  policy for optmizing the device placement is trained through repeated trials. Some of these works~\citep{dp_rl, Mirhoseini2018AHM_hdp,gdp_google} use a recurrent neural network or an attention network to predict a placement for each task and combine them into a placement of the whole graph.
Placeto~\citep{addanki2019placeto} uses graph embedding with RL and searches for better placements through a sequence of iterative placement improvements. It  is the closest prior work to ours and one of our comparison baselines (Section~\ref{sec:eval}). 

These RL methods for device placement have two limitations: 1) they consider a fixed device cluster; 2) the  cost of communicating between devices is not directly considered. As a result, they generalize poorly to new device clusters.

Many prior works have also focused on 
{specific types of systems. GiPH, instead, solves \emph{general} placement problems without domain-specific assumptions. For example, unlike fog/edge computing placement ~\citep{NAYERI2021103078_fogAI,salaht2020overview} with IoT~\citep{PALLEWATTA2022121_fog_iot,ddrl_iot} that assumes a layered or hierarchical network structure, GiPH can be applied to any network topologies. \citet{vne_gcn,vne_rl} solve the virtual network embedding problems assuming a relatively static physical network with a fixed number of devices, while GiPH allows topological changes and targets application contexts that are more dynamic (i.e.,  physical nodes can move and dynamically enter or exit the network). 

We consider the  objective of  \textbf{makespan (i.e., completion time) minimization}.
The Heterogeneous Earliest Finish Time (HEFT)~\citep{heft} scheduling algorithm is a highly efficient scheduling algorithm that 
combines task prioritization with a heuristic device placement strategy, and is another one of our baselines. More recent work \citep{Decima,Luo2021LearningTO} learns efficient online schedules with GNNs and RL. 
}


\section{Placement in Heterogeneous Computing}\label{sec:problem}
\begin{figure*}
    \centering
    \includegraphics[width=.85\linewidth]{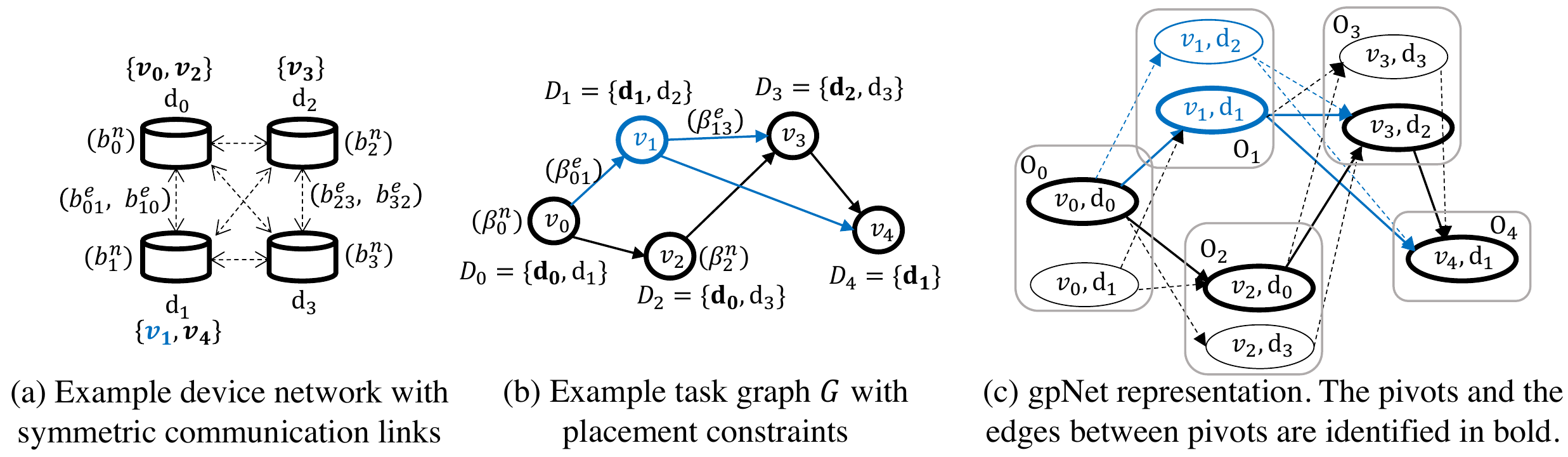}
    \caption{(a) Device network and (b) task graph of an example placement problem. The set of feasible devices for each task (placement constraints) is shown in (b). Node and edge features are in parentheses. (c) shows the gpNet representation for (a) and (b).}
    \label{fig:p1}
\end{figure*}

{
We consider a general placement problem in heterogeneous computing that aims to properly assign each task in a distributed application  to a device in a target computing network to optimize a performance criterion. We consider a distributed computing environment with 1) heterogeneous compute capabilities, 2) heterogeneous communication capabilities, and 3) placement constraints. }



A \textbf{target computing network} (e.g., Fig. \ref{fig:p1}(a)) consists of a cluster  $D=\{d_0,...,d_{m-1}\}$ of interconnected devices.  
 The compute devices in the system can be CPUs and/or GPUs that run at different speeds.  For example, a CNN can be much faster on a GPU than a CPU, and its actual running time can differ on different GPU modules.  Devices have \textit{different compute capabilities} (compute features) depending on the processor types, clock rates, etc.  We model the compute capability of a device $d_i$ by a vector of compute features $b^n_i$ (i.e., device compute speed, hardware type). There are also  \textit{placement constraints} resulting from hardware functionality and feasibility of devices, e.g., the task of acquiring LIDAR data can only be placed on a LIDAR sensor but not on a camera or an  edge server. 

We also consider a general \textit{heterogeneous communication network}. 
Devices can have different communication mechanisms (wired connection or wireless networks) to send data to each other. The specific means of communication defines the speed of data transmission, which can be different between pairs of devices. For example, a camera can be directly wired to a server that communicates wirelessly with mobile phones. We sometimes call a device cluster a device \emph{network} to emphasize the role of device connectivity in application performance. Each pair of devices $(d_i,d_j)$ has communication link features $b^e_{ij}$ (e.g., bandwidth, delay).  For the purpose of this paper, we assume the devices are fully connected and we only consider single-link paths (represented by edges) between devices. It is easy to generalize to more complex topologies by attaching very high communication losses to links that do not exist.

 A distributed application is defined by a directed acyclic \textbf{task graph} $G=(V, E)$, as shown in Fig. \ref{fig:p1}(b), where nodes $V=\{v_0,...,v_{n-1}\}$ represent computation tasks of the application and edges $E\subset V\times V$ represent inter-task data dependencies and communication. We define \emph{parents} of a task $v$ as those tasks on which $v$ has a data dependency, i.e., $\{u|(u,v)\in E\}$, and reversely $v$ is a \emph{child}  of its parent. Edges capture the precedence constraint that a child  can only start after it receives all data inputs from its parents.  The features of each node $v_i\in V$ and each edge $(v_i,v_j)\in E$,  represented by ${\beta}^n_i$ and ${\beta}^e_{ij}$, respectively, should be defined according to the optimization objective, e.g., to minimize the completion time, node and edge features may include the amount of compute of each task and data of  each  link. 


%



Given a distributed application $G=(V,E)$ and a device network $N$, a \textbf{placement} maps each  task in the application to a device in the network $\mathcal{M}: V\to D$. Each computation task $v_i$ can only be mapped to a subset of devices $D_i\subseteq D$ (Figs. \ref{fig:p1}(a) and (b)) due to placement constraints. The goal of the placement is to optimize a \emph{performance criterion} $\rho(\mathcal{M}|G,N)$ while satisfying $\mathcal{M}(v_i)\in D_i$ for all $v_i\in V$. $(G,N)$ defines a specific problem instance, and we denote a general placement as a triple $\mathcal{P}=(G,N,\mathcal{M}^{G\rightarrow N})$. 

We take performance to be the \textbf{makespan} (i.e., completion time) of an application, which is the time duration from the start of the first task's execution to the end of the last task's execution for a given application input. To simplify the problem, we assume the makespan is input-agnostic, e.g., CNN inference will have similar latency for any image.

Due to the data dependencies that exist among tasks, a child task can only start after it receives all the data from its parents. Therefore,  task execution must follow the precedence constraints defined by the  application DAG (i.e., a partial order). If we associate each node $i$ with a cost $c_i$, and  each edge $(i,j)$ with a cost $c_{ij}$, representing the computation and communication latency, respectively, the objective is 
\begin{equation*}
    \min_{\mathcal{M}}\rho(\mathcal{M}|G,N) =\min_{\mathcal{M}} \max_{p\in P(G)}\left(\sum_{i\in p}c_i + \sum_{(i,j)\in p}c_{ij}\right),
\end{equation*}
where $P(G)$ is the set of all paths from an entry node to an exit node in graph $G$, and $p$ is composed of all nodes and edges along the path. (The dependence of $c_i$ and $c_{ij}$ on $\mathcal{M}$ is omitted for simplicity.) Given a placement, the makespan is the total cost along the critical path (i.e., the path with the highest cost){, which can be determined in $\mathcal{O}(|E|+|V|)$ time by traversing the graph in topological order}.

{This placement problem is NP-hard, which means finding an exact optimal solution requires exhaustive search. A simpler version of the problem, which assumes unit-time computation and no communication delay, has been proven to be NP-complete~\citep{npcomp_scheduling}. Therefore, a simple one-step classification approach, where each node's ``class'' is the device on which it is placed, is not a viable solution. Like most prior work, GiPH tackles this challenge by considering task placement individually instead of attempting to determine the placement of the entire graph all at once, making the problem more tractable. }

\section{GiPH}\label{sec:giph}
\begin{figure}
    \centering
    \includegraphics[width=.8\linewidth]{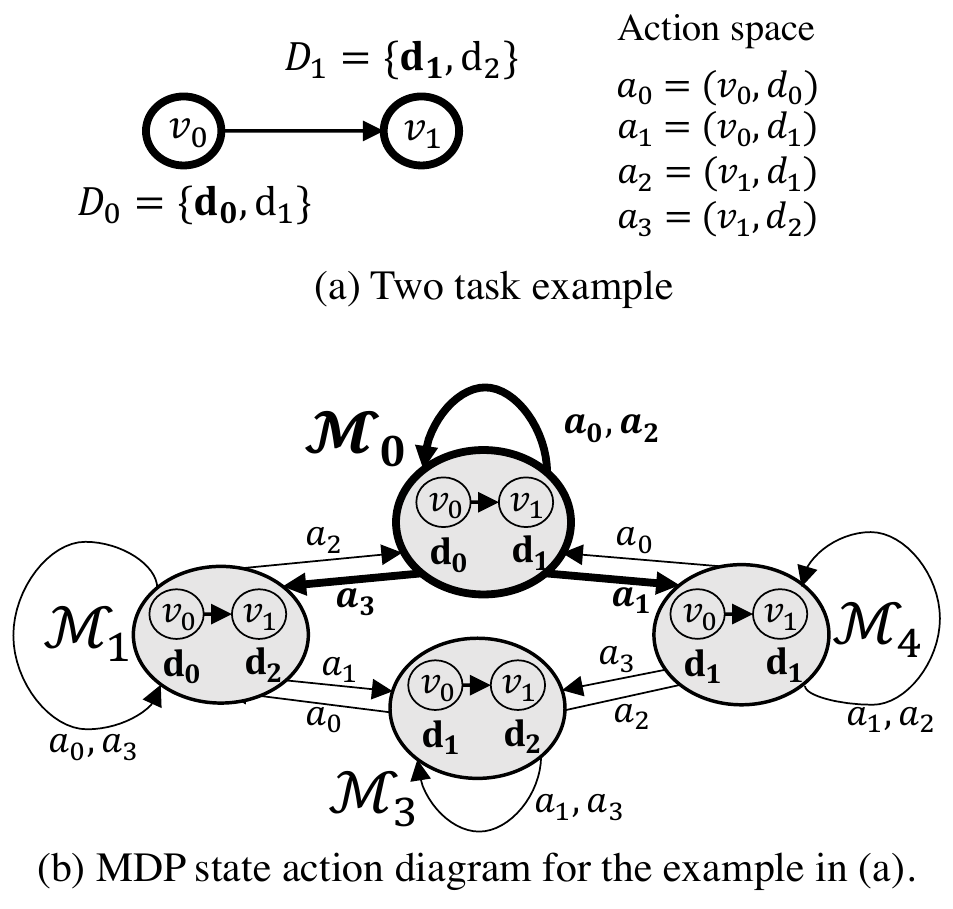}
    \caption{MDP (Markov decision process) of the placement search for a 2-task graph on 3 devices.}
    \label{fig:p2}
\end{figure}
This section introduces GiPH, an RL method that improves placement by relocating tasks. In \cref{sec:mdp}, we describe the formulation of the search problem and the associated Markov decision process (MDP) that underlies our RL approach. In \cref{sec:design}, we propose GiPH,  which utilizes a novel gpNet representation to encode task-level and device-level features, along with a scalable neural network design that summarizes the graph information and makes relocation decisions.

\subsection{Markov Decision Process Formalism}\label{sec:mdp}

We formulate the placement problem as a search problem, where given an initial placement, {a learned policy} iteratively {relocates some of the tasks}. Through making these incremental changes, the policy is able to search through the solution space and find better placements. Instead of trying to learn a policy that places the whole graph at once,  our search approach makes the learning simpler by only considering a small local search space at a time. {Focusing on incremental changes further allows us to generalize to arbitrary application task graphs and device clusters (\cref{sec:design}).} 

\paragraph{State:} Consider a single problem instance  $(G,N)$. For the search problem, we define the \textit{state space} as the set of all feasible placements $S_{G,N}=\{\mathcal{M}|\mathcal{M}(v_i)\in D_i,\forall v_i\in V\}$. (\textit{State} and \textit{placement} terms are  interchangeably used in the paper.) The size of the state space $|S_{G,N}|=\prod_{i=0}^{|V|-1} |D_i|$, since each task can be placed on any of the feasible devices for it. When there is no placement constraint { (i.e., $D_i=D$ for all tasks)}, $|S_{G,N}|=|D|^{|V|}$. For the two-task example shown in Fig.~\ref{fig:p2}(a), there are a total of 4 feasible placements, all shown as states in the transition diagram of Fig.~\ref{fig:p2}(b). 

\paragraph{Action:} {Observing that different placement configurations (states) differ in the placements of specific tasks, 
we define an \textit{action} to be a task and device pair $(v_i, d_j)\in V\times D$ that places  $v_i$ on device $d_j$. By including task selection in the action, we allow the policy to relocate tasks in any order and relocate for more than once. In this way, the search-based policy can go back and adjust the placement of the same task again after moving other tasks, exploring the state space. This is different from Placeto, which traverses each node in the graph only once and in an arbitrary order, impeding its ability to fully explore possible placement options.}


We only consider feasible actions $(v_i,d_j)$ such that $d_j\in D_i$. The size of the action space is thus $|A_{G,N}|=\sum_{i=0}^{|V|-1}|D_i|$, which is $|V||D|$ if there are no placement constraints. Fig. \ref{fig:p2}(b) lists all four actions for the simple two-task example and shows the deterministic state transition given an action taken at each state.
Note that the diameter, i.e., the length of the longest shortest path between any two states in the state transition diagram, is $|V|$ because one can always change from one placement to any other placement by moving each task node at most once. Therefore, even though the size of the state space grows exponentially with $|V|$, we can reach any state from any other state in $|V|$ steps. 

\paragraph{Reward:} The \textit{objective function} $\rho(\mathcal{M}|G,N)$ reflects how good a state $s=\mathcal{M}$ is. We assign intermediate reward  $r_t= \rho(s_{t+1}|G,N)-\rho(s_{t}|G,N)$, which mimics the advantage function \cite{rl} indicating the performance improvement after taking an action $a_t$ at a given state $s_t$. The goal of RL is to learn to take actions in order to maximize the  expected return $\sum_{t=0}^{T}\gamma^tr_t$, where $T$ is the  episode length. When $\gamma=1$, the expected return is the expected performance improvement between the final state $s_{T}$ and an initialized state $s_0$, i.e., the policy tries to maximize $\mathbb{E}[\rho(s_{T}|G,N)] - \mathbb{E}[\rho(s_{0}|G,N)]$. When the distribution of the initial placement is fixed,  the latter term is constant for a $(G,N)$ pair 
and RL is effectively improving the expected performance of the final placement through maximizing the expected return. When $\gamma < 1$, the policy seeks more immediate reward as future rewards are discounted. In this case, the policy learns to search more efficiently {(i.e., increase the reward the most)} at the beginning of the RL episodes.   



\subsection{GiPH Framework}\label{sec:design}

\begin{figure*}
    \centering
    \includegraphics[width=.9\linewidth]{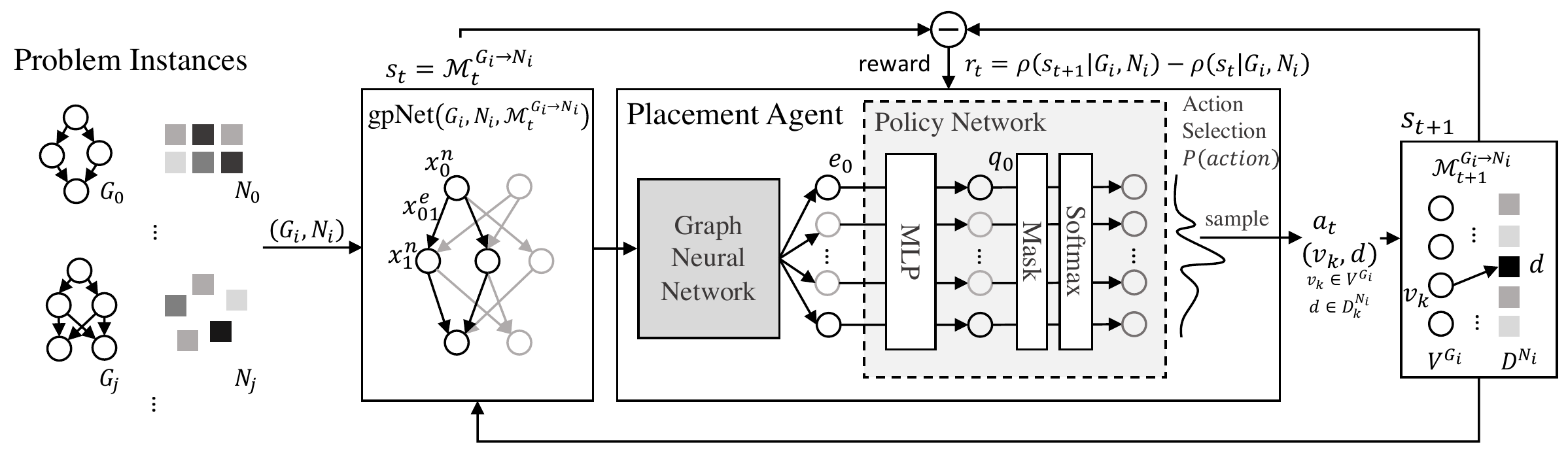}
    \caption{GiPH neural network design. A given problem instance $(G_i, N_i)$ is first transformed into a graphical gpNet representation, which is then passed through a GNN to generate feature embeddings for each feasible task-device placement pair. The policy network then chooses an action (an incremental task-device re-placement) by estimating its performance improvement.}
    \label{fig:p3}
\end{figure*}

By leveraging the MDP, GiPH can learn to iteratively optimize a placement by relocating tasks. The framework of GiPH is shown in Fig. \ref{fig:p3}. Given a placement problem of an arbitrary task graph $G_i$ and target network $N_i$, the placement agent starts a search  from an initial placement $\mathcal{M}^{G_i\rightarrow N_i}_0$. {This initial placement can be generated using some simple strategies, or it can be a placement that requires improvement.} Following the MDP, at each step, the agent  takes as  input the current state of the search, $s_t=\mathcal{M}^{G_i\rightarrow N_i}_t$, decides a task relocation step $a_t\in A_{G_i,N_i}$ that modifies the current placement to $s_{t+1}$, and observes the improvement of the objective $\rho$ as the reward $r_t$. 

{GiPH comprises three key components.  Firstly, the novel graph representation gpNet  (\cref{sec:representation}) explicitly encodes both task-level and device-level features of the current placement state. This facilitates the learning of a fully generalizable policy that can be applied to different placement problems. Secondly, a graph neural network (\cref{sec:gnn}) computes an embedding for each action (i.e., a task relocation step) based on the gpNet representation. The GNN leverages the graph structure of the placement problem to summarize relevant information and generate high-quality embeddings. Finally, a policy network (\cref{sec:policy}) uses a score function to make decisions on which action to take. The GNN and the policy network are jointly trained to optimize the placement policy.  }

\subsubsection{gpNet Representation}\label{sec:representation}
We have formulated a discrete MDP for the placement search problem given $(G,N)$. For the learned policy to be fully generalizable across different problem instances, we need a representation for a general placement $\mathcal{P}=(G,N,\mathcal{M}^{G\rightarrow N})$. This representation must capture the compute and communication requirements of the task graph $G$ and the  compute and communication capabilities of the device network $N$, enabling the policy to learn the  performance function $\rho$ given $\mathcal{P}$. This general representation enables the learned policy to be applied to different placement problems without being specific to a particular task graph or device network, improving the policy's ability to generalize and achieve good performance in various scenarios.


To this end, we present \textbf{gpNet}, a novel and universal graph representation of the placement that encapsulates features of both the task graphs and the device networks with placement constraints. gpNet generates a unique graph $H=(V_H,E_H)$ given a general placement $\mathcal{P}=(G,N,\mathcal{M}^{G\rightarrow N})$, where $G=(V,E)$ is an arbitrary task graph with node features $\bm{\beta}^n$ and edge features $\bm{\beta}^e$, and $N$ is an arbitrary device network composed of a device cluster $D$ with device compute features $\bm{b}^n$ and communication link features $\bm{b}^e$. $H\leftarrow gpNet\left(G=(V,E,\bm{\beta}^n,\bm{\beta}^e),N=(D,\bm{b}^n,\bm{b}^e),\mathcal{M}^{G\rightarrow N}\right)$.

\paragraph{Node generation:} $V_H=\{(v_i,d_j)| d_j\in D_i \textrm{ for } i=0,...,|V|-1\}$. Each node in $H$ represents a feasible placement of $v_i\in V$ on device $d_j\in D_i$, and is labeled $(v_i,d_j)$, e.g., Fig.~\ref{fig:p1}(c)
. The node features $x^n_u$ of a node $u=(v_i,d_j)$ are a function, $f_n$, of the task features $\beta^n_i$ and the device features $b^n_j$, i.e., $x^n_u=f_n(\beta^n_i, b^n_j)$. The set of nodes for all possible placements of a task $v_i$ forms a group of placement options  $O_i=\{(v_i,\cdot) \}\subseteq V_H$. Nodes whose labels are in the current placement $\mathcal{M}^{G\rightarrow N}$ are called \textit{pivots} and form a set $V_{H,\mathcal{P}}\subseteq V_H$. The subgraph induced by $V_{H,\mathcal{P}}$ thus contains all information about the current placement $\mathcal{M}^{G\rightarrow N}$. Non-pivot nodes, on the other hand, represent a potential task \emph{re-placement}.  Each node in $H$ also corresponds to one action defined in the search problem described in \cref{sec:mdp}. 

\paragraph{Edge generation:} We add edge $(u_1,u_2)$ with $u_1=(v_i,d_k), u_2=(v_j,d_l)$ to the gpNet  $H$ if $(v_i,v_j)\in E$ and at least one of $u_1$ and $u_2$ is a pivot, i.e., $u_1\in V_{H,\mathcal{P}}\textrm{ or }u_2\in V_{H,\mathcal{P}}$.  In this way,  each non-pivot  node $(v_i,d_j)$  only has edges pointing to or from pivots that contain the current placement information of its parents and children. For example, in Fig.~\ref{fig:p1}(c), $(v_1,d_2)$ has an incoming edge from $(v_0,d_0)$, which contains the {current} placement information of $v_1$'s parent task $v_0$,  and outgoing edges to $(v_3,d_2)$ and $(v_4,d_1)$, which contain the {current} placement information of $v_1$'s child tasks $v_3$ and $v_4$.  Thus,  \emph{non-pivot node $(v_i,d_j)$ has a local graph structure corresponding to $v_i$ being re-placed to $d_j$}. The edge feature $x^e_{u_1u_2}$ of $(u_1=(v_i,d_k), u_2=(v_j,d_l))$  is a function, $f_e$, of the data link features $\beta^e_{ij}$ and the communication link features $b^e_{kl}$. 
The resulting graph $H$ has $|V_H|=\sum_{i=0}^{|V|-1}|D_i|$ nodes and $|E_H|=\sum_{i=0}^{|V|-1}\left(|D_i||E_i|\right) - |E|$ edges, where $|E_i|$ is the degree of $v_i$ in $G$. Both $f_n$ and $f_e$ can be any functions that combine the features of the task graph and the device network (e.g., concatenation).

{See Appendix \ref{alg:gpnet} for the full algorithm.} Our proposed gpNet unifies a given application graph $G$ and device cluster $N$ into a single graph that \emph{captures all device- and task-related features} for making a placement update decision. The original task dependencies in $G$ and placement constraints are implicitly present in the output gpNet $H$ by construction.  gpNet also generalizes to different problem instances: we can construct a gpNet for any placement of an arbitrary task graph-device network problem pair $(G,N)$.

\subsubsection{Scalable and Generalizable Graph Embedding}\label{sec:gnn}
 GiPH must first convert the placement information, represented in graphical form by $gpNet(G_i,N_i,\mathcal{M}^{G_i\rightarrow N_i}_t)$, into features that can be passed to the policy network. 
 Creating a flat vector representation is not scalable because it cannot handle graphs of arbitrary sizes and shapes (which depend on the specific task graph, target network and constraints). 
 
 GiPH achieves scalability using a graph neural network (GNN) \citep{learn_graph,Battaglia2018RelationalIB} that embeds the state information in a set of embedding vectors. Taking a gpNet as input with node features $\bm{x}^n$ and edge features $\bm{x}^e$ composed as described in \cref{sec:representation}, GiPH propagates information in a sequence of message passing
 \begin{equation}\label{eq:embedding}
     e_u=h_2\left(\sum_{v\in \xi(u)}h_1\left([e_v\mathbin\Vert x_{vu}^e]\right)\right) + x_u^n,
 \end{equation}
 for each node $u$ in the gpNet, where $\xi(u)$ is the set of parents of $u$, who have aggregated messages from all of their parents. $h_1(\cdot)$ and $h_2(\cdot)$  are non-linear transformations over vector inputs with trainable parameters. The message passing is done in both forward and backward directions  with separate parameters, each summarizing information about the subgraph of nodes that can be reached 
 {from  $u$ and nodes that can reach $u$}.
GiPH concatenates the two summaries along each direction as the node embeddings. For a node with label $(v,d)$, this embedding thus captures the local placement information if $v$ is placed on $d$ (i.e., if an action $(v,d)$ is taken). 
 Adopting a GNN also helps generalizability because it automatically learns high-level features that are statistically important through end-to-end training, and the model learned can generalize (and scale) to unseen graphs.

\subsubsection{Policy Network and Actions}\label{sec:policy} The policy network consists of a multi-layer perceptron (MLP), an optional mask layer, and a softmax layer (Fig. \ref{fig:p3}). We use the per-node embedding from the GNN to compute a score $q_a=g(e_a)$ for each action $a$ in the action space $ A_{G_i,N_i}$ (represented as nodes in the gpNet). $g(\cdot)$ is a score function implemented as a MLP that computes a scalar value for an embedding vector. The score $q_a$ quantifies how good an action is given the current state $s$. GiPH then uses a softmax layer to output a probability of selecting each action based on the score $P(a|s)={\exp{(q_a)}}/{\sum_{b\in A_{G_i,N_i}}\exp{(q_b)}}$. An optional mask layer can be placed before the softmax  to mask out undesired actions. The final output is a probability distribution over all feasible actions.

Since we may have input gpNets of arbitrary sizes, instead of using a fixed-size policy network, the GiPH policy network  adopts a score function to evaluate individual actions {(i.e., nodes in gpNet), based on their local graph connectivity and encoded placement information}. Thus, it \emph{does not depend on the size of gpNet}. 

 To improve the sample efficiency and force exploration, we mask out actions that do not change the current placement (e.g., $a_0,a_1$ at state $\mathcal{M}_0$ in the example Fig. \ref{fig:p2}(b)) because no new information will be acquired by taking those actions. We also mask out actions that will result in moving the same task consecutively twice because 
 {we expect the policy to find a better device for a task within one move}. 
\section{Evaluation}\label{sec:eval}
In this section, we evaluate the performance of our proposed GiPH for makespan minimization. We assume a heterogeneous computing environment where the computation time and communication time can be estimated from compute (task) and communication (data link) features of the device network. {We first evaluate GiPH using synthetic data with randomly generated task graphs and device networks that cover a wide range of cases in \cref{sec: synthetic}, {and specifically test on deep learning graphs in \cref{sec:dl-eval}. }In \cref{sec: cav} we present a case study of applying GiPH to a realistic setting of autonomous intersection management for Connected Autonomous Vehicles (CAVs) using cooperative sensor fusion.} 

We compare GiPH with the following baseline algorithms:
\begin{itemize}[leftmargin=*]
    \item \textbf{Random placement sampling}: generating random placements of the task graph by sampling a feasible placement for each task from a uniform random distribution. This random baseline is representative of the average placement ``quality'' {without GiPH's intelligent search}.
    \item \textbf{HEFT} ~\citep{heft}: {the state-of-the-art} heuristic scheduling algorithm for heterogeneous computing {that we use as benchmark}. It prioritizes tasks with high-level features and allocates each task, in order of priority, to a device that finishes the task the earliest ({Earliest Finish Time (EFT)} device selection).
    \item \textbf{Random task selection + EFT device selection}: a heuristic of placement search, where at each step a task in the graph is randomly selected and placed according to EFT. {It is a direct adaption of  HEFT as a search-based policy.}
    \item \textbf{GiPH task selection + EFT device selection}: the version of GiPH without gpNet {(ablation study).} At each step, instead of deciding a task-device pair, the RL agent only selects a task. The task is then placed according to EFT. 
    \item \textbf{Placeto}~\citep{addanki2019placeto}: a search-based device placement algorithm that is the closest {prior work to ours}. It 
    {also does incremental placement, but does not consider re-placing the same task or device network features. }
    \item {\textbf{RNN-based placer}: another RL baseline based on the hierarchical model for device placement (HDP)~\cite{Mirhoseini2018AHM_hdp}. The Placer traverses the graph in topological order and directly decides the device assignment of each operator through an RNN-based policy network.} 
\end{itemize}

\paragraph{Evaluation metrics:} 
We evaluate each algorithm's \textit{placement quality} and \textit{adaptivity}.
We evaluate the placement quality through the completion time of a task graph, i.e., makespan. Since the makespan can vary significantly  on different problem instances, we follow \citet{heft} in normalizing the makespan to an {instance-dependent} lower bound, defining the Schedule Length Ratio: 
\begin{equation*}
    SLR = \frac{makespan}{\sum_{v_i\in CP_{MIN}}\min_{d_j\in D_i}{w_{i,j}}},
\end{equation*}
where $w_{i,j}$ is the expected time of running task $v_i$ on device $d_j$ and $CP_{MIN}$ is the critical path based on the minimum computation cost of each task node. 
The placement algorithm that gives the \emph{lowest} SLR is the best with respect to the placement performance. We report the average SLR of different problem instances. 
%
We evaluate the adaptivity of the algorithms by  measuring the average SLR achieved after each network change following a random device addition and deletion procedure. 
\edited{}
{\paragraph{Experiment details:} 
Both GiPH and Placeto use a two-layer feed-forward neural network (FNN) with the same number of hidden units as the input dimension for node and/or edge feature pre-embedding {before message passing}. 
Messages are aggregated by mean. GNN implementations of GiPH and Placeto have comparable sizes, and are detailed in Appendix \ref{sec:training_detail}  (e.g., features, neural network sizes, running time). Both are trained using  the  policy gradient method REINFORCE~\citep{reinforce}  with 200  episodes.  


For the RNN-based placer network, we follow the HDP paper and use a sequence-to-sequence model with a bi-LSTM for the encoder and a unidirectional LSTM with an attention mechanism for the decoder. 
Since HDP does not aim to generalize to new device networks or new application graphs, we only compare the placement quality by training a new Placer policy on each test case with 4 Placer samples each time until the latency is no longer improved. 

All policies are trained using Adam optimizer with a fixed learning rate $0.01$. ReLU activation is used. The discounting factor $\gamma=0.97$. The placement performance is  evaluated using a runtime simulator (execution model and latency model detailed in Appendix \ref{sec:simulation}).
}

We separately generate training and test datasets, each composed of a set of task graphs and a set of device networks.  For testing, all search-based policies start from the same initial placement for fair comparison. Since the action space grows linearly with the number of computational tasks, we set the  episode length to be  multiples of the number of tasks in the task graph, and empirically find that twice the size of the graph $2|V|$ step are enough  for the policy to converge to a solution. Since Placeto fixes the number of search steps to $|V|$, we start a new search episode for Placeto after $|V|$ steps.  Each policy outputs the SLR of the best placement found so far within the episode.  
\begin{figure*}
    \centering
    \includegraphics[width=\linewidth]{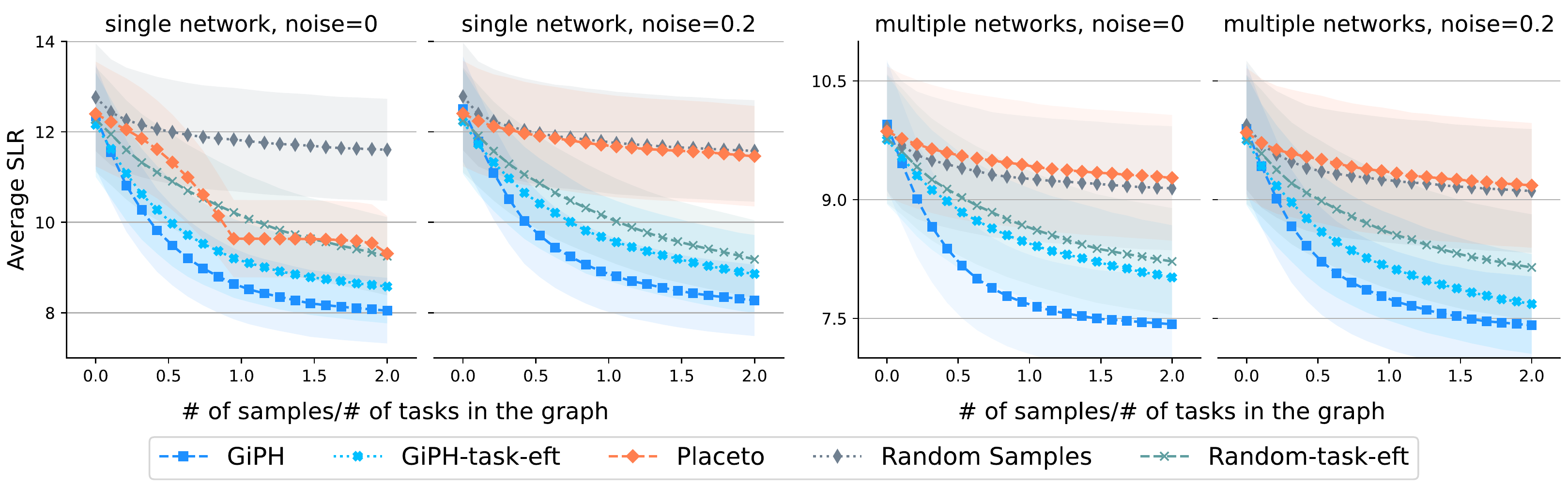}
    \caption{Placement quality and search efficiency of search-based policies. Noise=$0.2$ means the communication and computation time can vary up to $20\%$ of the average value. GiPH consistently finds placements with the lowest SLR (i.e., best performance). 
    }
    \label{fig:sampling}
\end{figure*}
\begin{figure*}[t]
    \centering
    \vskip 0.1in
    \includegraphics[width=\linewidth]{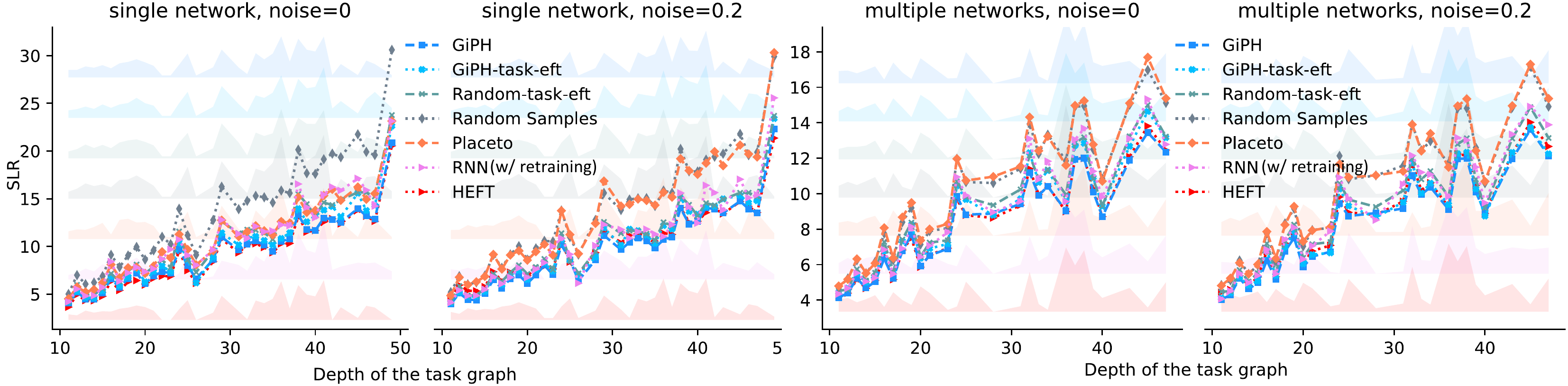}
    \caption{Average SLR with respect to the depth of the task graph. The background colorbars show the standard deviation of SLR for each method in the order they are listed in the legend. }
    \label{fig:single_depth}
\end{figure*}

\subsection{General Task Graphs and Networks}\label{sec: synthetic} 
\paragraph{Dataset:} We follow the parametric method used by ~\citet{heft} to generate random task graphs and random device networks with various characteristics (e.g., number of devices, average delay of the device networks, etc.) depending on the input parameters. 
To evaluate our work on a full range of task graphs and device networks, we decouple their parameters so that each can be generated independently. The data generation process is detailed in Appendix \ref{sec:data}. 
Our generators also {randomly} specify placement constraints by  adding a 
hardware-requirement property to each task and a corresponding hardware-support property to each device. {One input parameter specifies  the average number of feasible devices for each task.}




{\paragraph{Generalizability:} We consider two cases: (1) The \textit{single-device-network case}, where the search-based methods are trained and tested on a single  device network. This case mainly considers application-level generalization, and is the problem setting used in prior work, including Placeto.  
A set of 300 randomly sampled task graphs is split equally for training and testing.  (2) The \textit{multiple-device-network case}, where multiple device networks, with varying compute and communication capacities per device, are used for training and testing. This case further includes device-network generalization. $500$ test cases are sampled from {combinations of} $10$ device networks and $120$  graphs. All experiments are done with and without $20\%$ noise added to the computation and communication times, which can model the random performance of real systems {and estimation errors}.  

The average SLR across test cases as a function of the number of the search steps is shown in Fig.~\ref{fig:sampling}. In all cases, our GiPH policy outperforms other search policies and more rapidly finds better placements within fewer search steps. It achieves up to $30.4\%$ lower completion time compared to the random baseline, which represents the average placement ``quality''.  GiPH also exhibits resistance to variations in the communication and computation times (noise). GiPH-task-EFT,  without using  gpNet, is not as good as GiPH, but the RL on the task selection still allows it to outperform the random selection under the Random-task-EFT policy. In contrast, the performance of the Placeto policy significantly degrades under noise, probably because the  agent cannot de-couple the noise sources without a proper representation of the device network. When multiple networks are involved, Placeto even becomes worse than random  because, without considering device-level features, the policy learns false local optima that no longer exist in a new device network. 

 Fig.~\ref{fig:single_depth} shows the SLR of the final placements found by different algorithms on the testset with respect to the depth of the task graph. {As the task graph grows in depth, the SLRs for all methods increase because the critical paths are longer, increasing the makespan of the task graphs.} GiPH outperforms other search-based methods in most of the cases and is comparable to the state-of-the-art HEFT. 
}

\begin{figure}
    \centering
    \includegraphics[width=\linewidth]{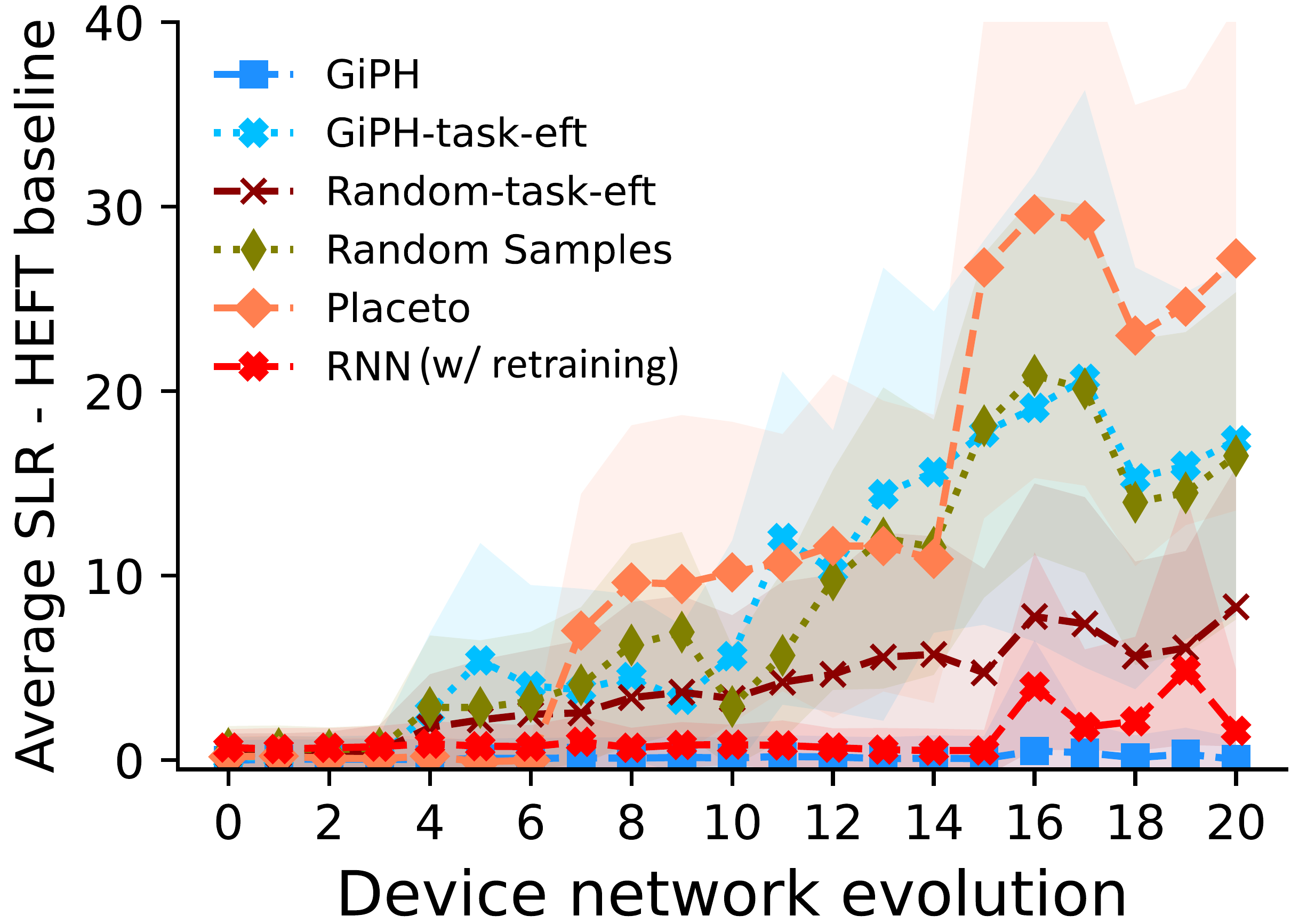}
    \caption{Adaptivity to device network changes. GiPH maintains stable performance, while other search methods achieve worse SLR as the network changes more.}
    \label{fig:generalizability}
\end{figure}

\paragraph{Adaptivity:}
We evaluate the performance of the learned policies on a changing device network. {The network initially has 20 devices, and as the network evolves, some of the devices are randomly removed and later replaced with new devices of lower capacities ({i.e.,} higher cost). The total number of devices is between 16 and 20.} {This may model devices running out of battery over time; to conserve energy, they may reduce their compute and communication resource expenditures.} Fig.~\ref{fig:generalizability} shows the average SLRs of different algorithms compared to the HEFT baseline, calculated across 20 different application graphs running on the contrived device network.  As the number of high-cost devices grows, the SLR for random samples increases due to higher average compute and communication times. Placeto again performs worse than random, being unable to adjust its policy as the device cluster changes. GiPH-task-eft fails to adapt because, without using  gpNet to encode information about alternative placements, the policy cannot correctly decide which task to relocate next. 
{Another baseline is the RNN-based placer, whose policy is retrained every time the network changes.  With significant retraining, it maintains a low SLR value. However, GiPH learns a fully generalizable policy, and}  is the only search-based method that maintains stable performance (with almost the same SLR as HEFT) and adapts well to the changes in the device network without retraining.

\begin{figure}
    \centering
    \includegraphics[width=\linewidth]{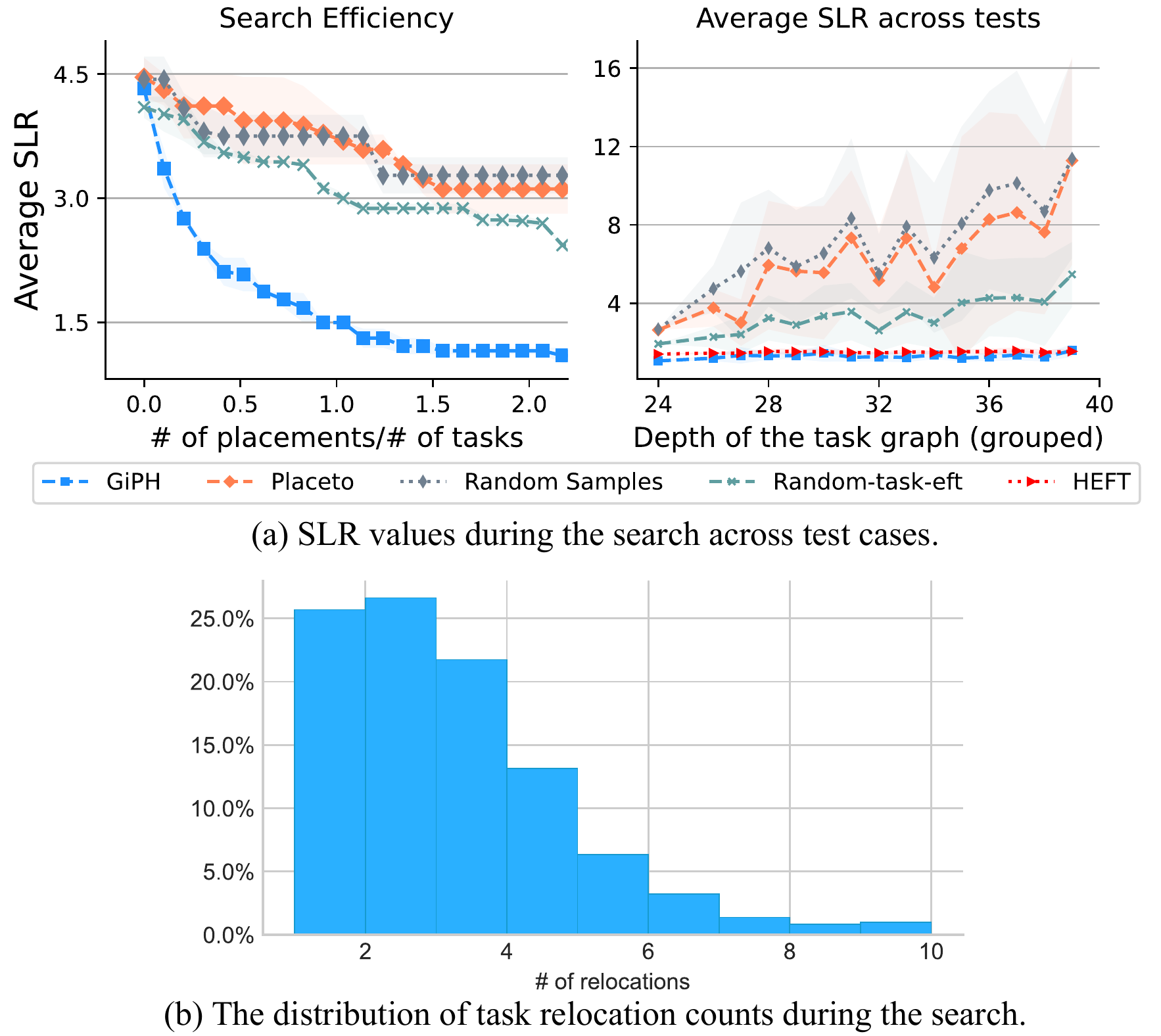}
    \caption{(a) Performance evaluation on deep learning computation graphs. (b) Counts distribution (only considering non-zero counts), with some tasks being relocated more frequently than others.}
    \label{fig:dl}
\end{figure}

{
\paragraph{Alternative implementations:} Our GNN design has two key aspects: (1) It fully incorporates per-edge features.
(2) It takes into account the partial ordering that defines the acyclic dependency of computation, by constructing message passing in both forward and backward directions with separate parameters. We further evaluate GiPH (1) without GNN, (2) without per-edge features, (3) with uni-directional GraphSAGE layers, (4) with k-step two-way message passing, and (5) without gpNet.  Our proposed GNN design shows better convergence properties (Appendix \ref{sec:ablation}).

{
\subsection{Deep Learning Graphs} \label{sec:dl-eval}
We specifically look at deep learning (DL) applications and evaluate GiPH on DL computation graphs generated by ENAS~\citep{enas}, an automatic model design approach. A dataset of 300 graphs, using the recurrent cell designs  from ENAS on the Penn Treebank benchmark for language models (details in Appendix \ref{sec:dl-graphs}), is split equally for training and testing. Each graph contains 200-300 operators. Similar to~\citet{addanki2019placeto,Mirhoseini2018AHM_hdp}, we partition the operators into predetermined groups and place operators from the same group on the same device. The grouping is done by iteratively merging the operator with in-degree one and lowest cost into its sole predecessor until the graph size is reduced to 40 nodes. 

Training and testing are both done on one single simulated device network containing 8 devices. The evaluation results are shown in Fig. \ref{fig:dl}(a), where all search-based policies start from the same initial placement for fair comparison. GiPH again outperforms all the baselines by selectively relocating specific task groups based on their current placement, resulting in improved search efficiency compared to Placeto. Placeto  traverses the graph and visits all nodes equally, whereas GiPH can adjust the placement of ``critical'' nodes more frequently within the same number of search steps. Figure \ref{fig:dl}(b) depicts the frequency distribution of relocation counts during the search using GiPH.

}

\subsection{Case Study: Cooperative Sensor Fusion}\label{sec: cav}

We  present a
{n experimental} case study of applying GiPH to a realistic setting of autonomous
 intersection management in CAVs. We consider an autonomous driving scenario where Roadside Units (RSUs) at major intersections gather real-time image data from connected infrastructure camera sensors (ISs) and camera and LIDAR data from CAVs around the intersection to plan the CAV trajectories  in a timely manner. The data collection, communication, and processing form a time-sensitive dataflow task graph whose feasible placements need to be rapidly determined as CAVs move and their communication costs change.  

\paragraph{Realistic application traces:} We use a traffic simulation tool Simulation of Urban MObility (SUMO)~\citep{SUMO2018}  to simulate traffic  within a 6-block area in the center of Tempe AZ (Figure \ref{fig:sumo}(a)).  Each major intersection is equipped with one RSU and four CISs, and  will ``interact'' with CAVs if they are within $400$m of the RSU. We consider  the sensor fusion pipelines in \citet{andert2022accurate} for sensing and localization, which involve camera object detection, LIDAR object detection, and fusion of the processed data for localization. The first two tasks need to run on  GPUs, and the last one can run on any compute devices (CPU). 
We simulate the traffic with 3980 vehicles generated in an hour period ($10\%$ of CAVs) and collect application traces at 10 second intervals (Figure \ref{fig:sumo}(b) as an example). 

\begin{table}
\caption{Running time measurements of each task on device types A, B, and C with standard deviation values (in milliseconds)}
\label{tab:measurements}
\begin{center}
\begin{small}
\begin{sc}
\begin{tabular}{lccc}
\toprule
&  \makecell{Type A
}  & \makecell{Type B
}  & \makecell{Type C
} \\
\midrule
\makecell{Camera
}& 53$\pm${22} & 36$\pm${8} & 9$\pm${4}\\\hline
\makecell{LIDAR
}& 14$\pm${3} & 7$\pm${3} & 3$\pm${2}\\\hline
\makecell{CAV Data Fusion
}    & 35$\pm${9} & 35$\pm${4} & 11$\pm${9}\\\hline
\makecell{RSU Data Fusion
}    & 250$\pm${430} & 250$\pm${370} & 28$\pm${22}\\
\bottomrule
\end{tabular}
\end{sc}
\end{small}
\end{center}
\vskip -0.1in
\end{table}
\begin{figure}
     \centering
     \includegraphics[width=.9\linewidth]{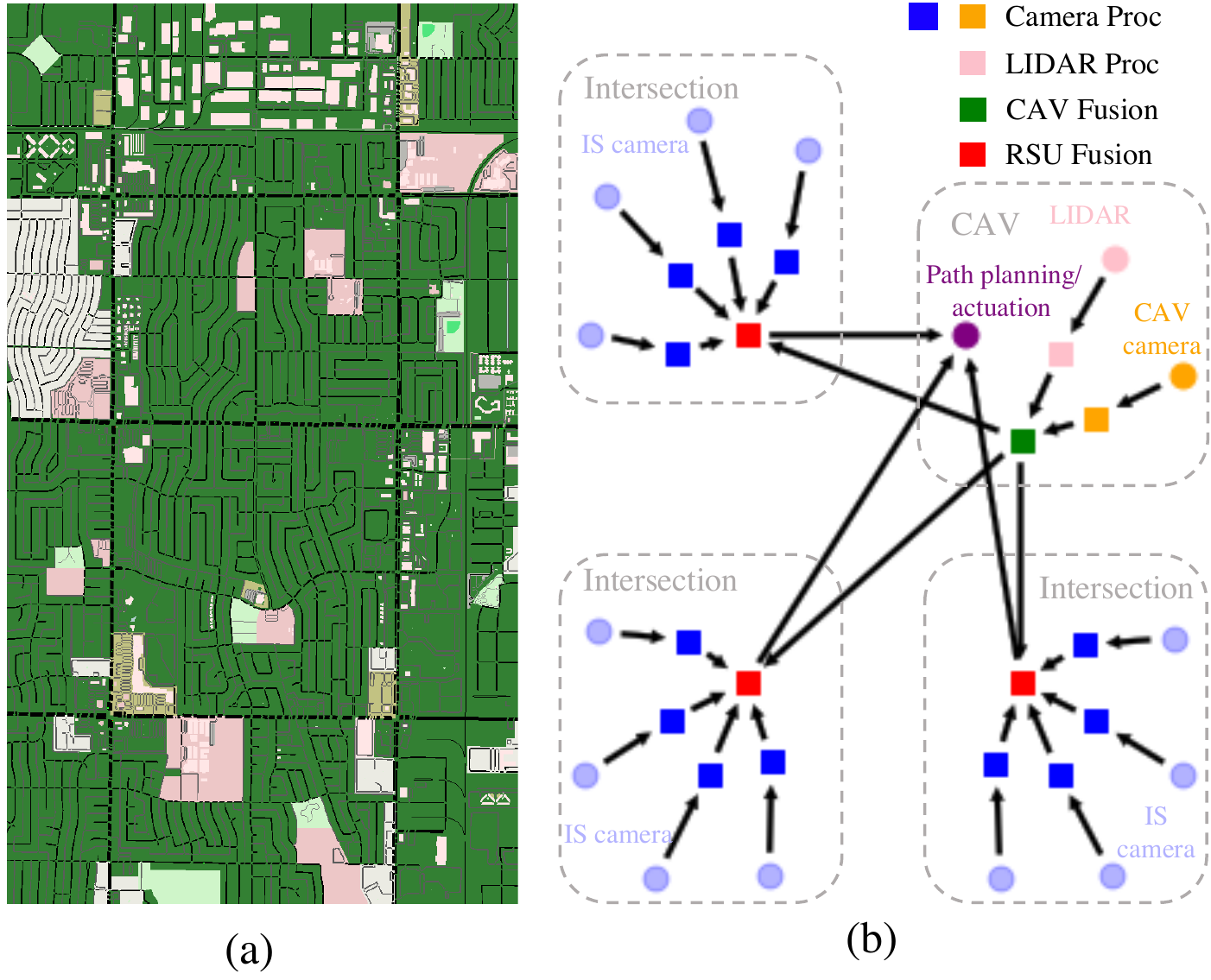}
     \caption{(a): SUMO traffic simulation area. (b): A snapshot of the application graph of autonomous traffic management for a CAV between three intersections.  }
     \label{fig:sumo}
 \end{figure}
 
 \begin{figure}
    \centering
    \includegraphics[width=\linewidth]{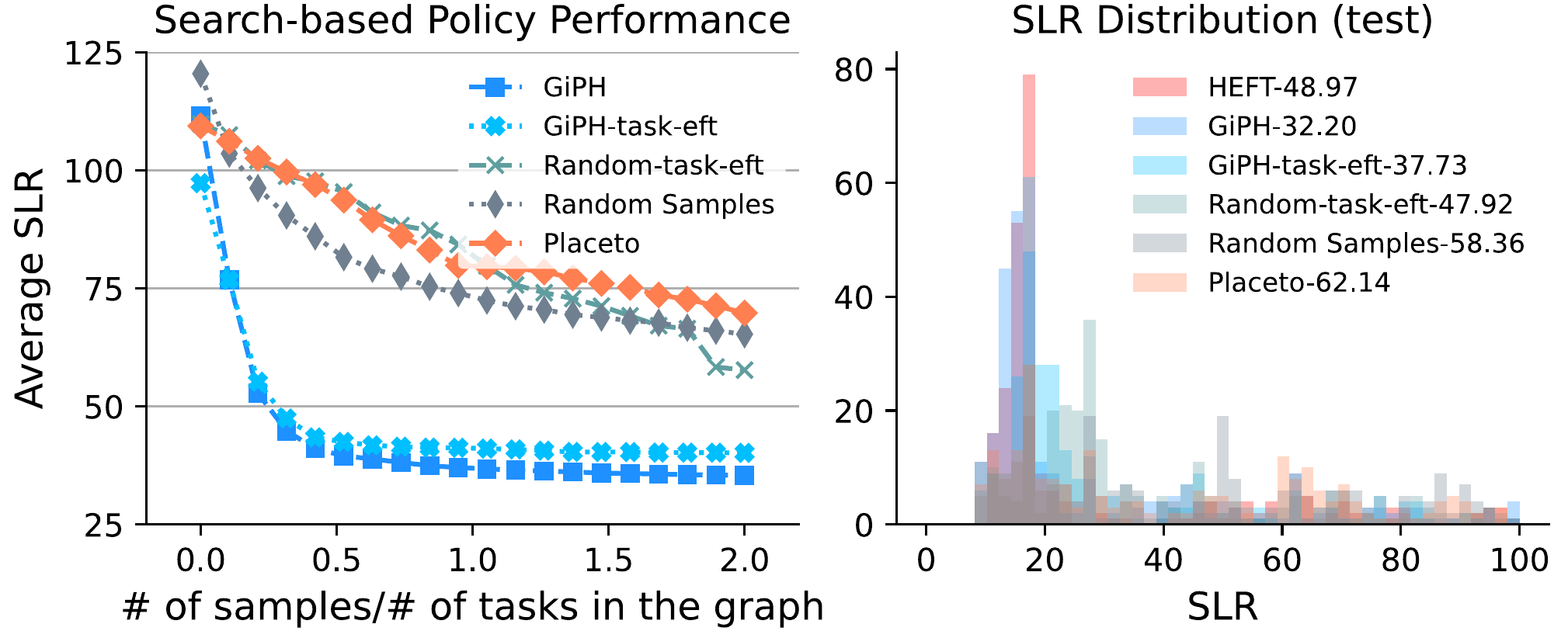}
    \caption{Case study using the application traces for autonomous intersection traffic management. (a) Search efficiency of search-based policies. (b) SLR distribution (under 100) with mean values included in the legend. Data distribution above 100 is very sparse and not shown in the figure for visualization. }
    \label{fig:realistic}
\end{figure}

\begin{figure}
    \centering
    \includegraphics[width=.9\linewidth]{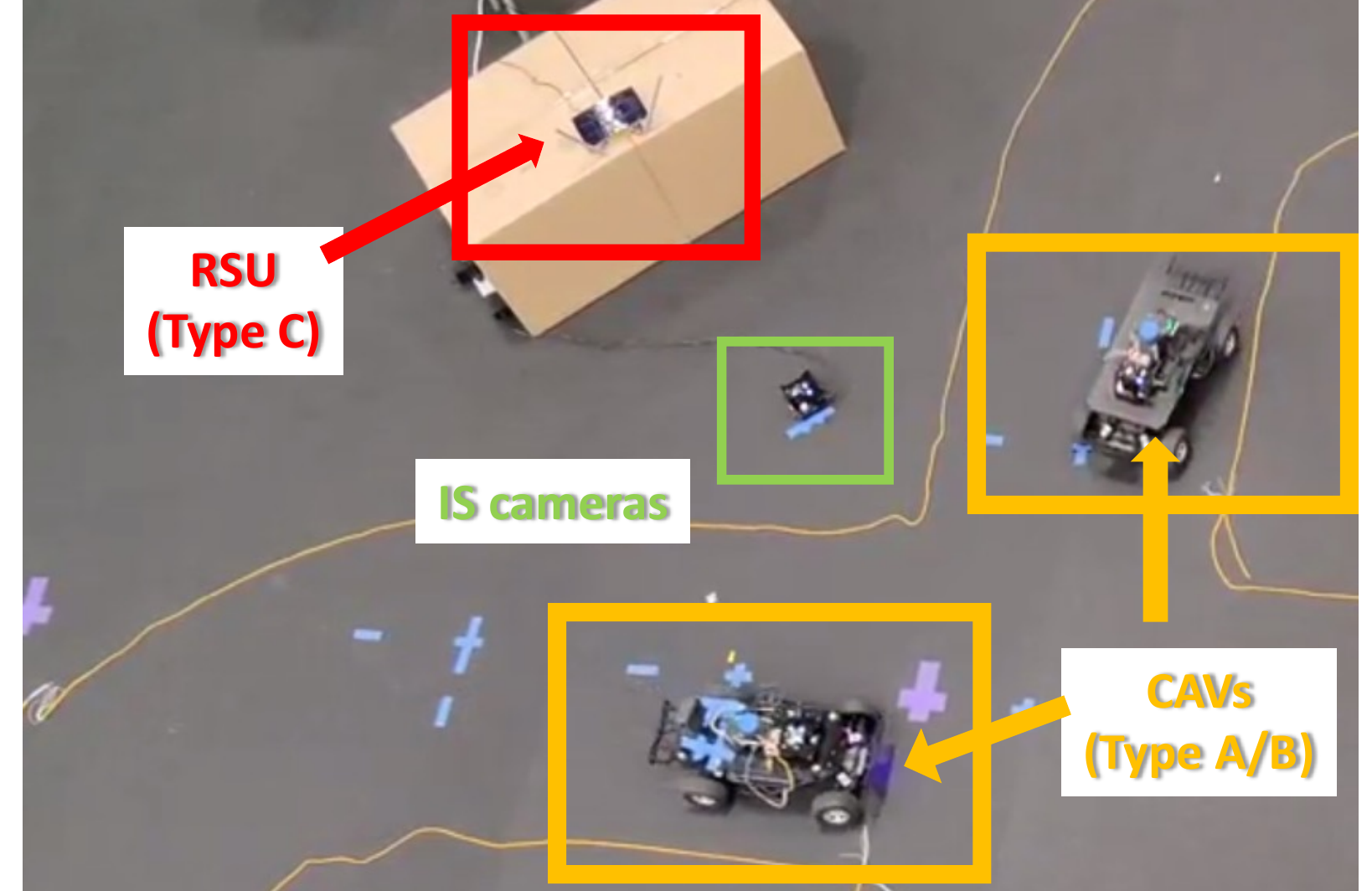}
    \caption{Real-world deployment of the sensor fusion pipeline (source: \citet{andert2022accurate})}
    \label{fig:deployment}
\end{figure}
 
\begin{table*}
\caption{Measurements of the relocation overhead of each task in a small-scale deployment. }
\label{tab:deployment}
\begin{center}
\begin{small}
\begin{sc}
\begin{tabular}{lcccc}
\toprule
&  \makecell{Data migration\\ (bytes)
}  & \makecell{Static initialization\\data (kilobytes)
}  & \makecell{Startup time\\ Type A (ms)
}  & \makecell{Startup time\\ Type C (ms)
} \\
\midrule
\makecell{Camera
}& 11494 & 72173.525 & 4273.73 & 794.66\\\hline
\makecell{LIDAR
}& 560 & 24.576 & 60.98 & 9.26\\\hline
\makecell{CAV Data Fusion
}    & 11796 & 38.110 & 0.39 & 0.11\\\hline
\makecell{RSU Data Fusion
}    & 20907 & 38.950 & 2.83 & 1.00\\
\bottomrule
\end{tabular}
\end{sc}
\end{small}
\end{center}
\vskip -0.1in
\end{table*}

\paragraph{Placement experiment:} We first establish a realistic latency model by measuring the actual running time of each task on Jetson Nano (Type A), Jetson TX2 (Type B), and Core i7 7700K with GTX1080 (Type C). The measurements are summarized in Table \ref{tab:measurements}.

\begin{figure}
    \centering
    \includegraphics[width=\linewidth]{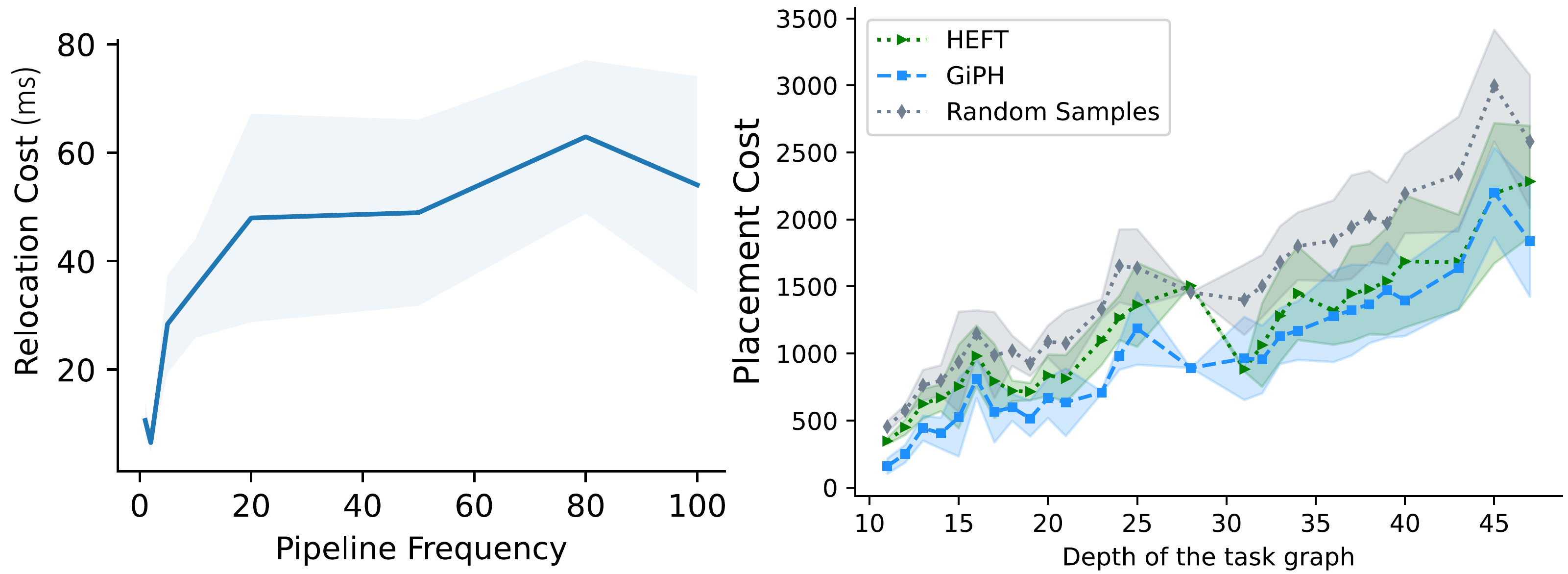}
    \caption{Left: incurred relocation cost with respect to pipeline frequency. Right: Using the placement found by GiPH, HEFT, and Random policy, the total energy cost across test cases. }
    \label{fig:rcost}
\end{figure}
There are 36 RSUs located in the major intersections. We model them as Type-C devices. To introduce alternative placement options for running the sensor fusion pipelines, we randomly place 40 edge compute devices (10 type-A, 10 type-B, and 20 type-C devices) in the simulated area that can serve as additional RSUs for sensor fusion and trajectory planning. Finally, we model the communication network with a data rate that decays exponentially with distance. Details are included in the Appendix \ref{sec:case_study}.

We  evaluate GiPH and other search-based policies on over 900 placement cases that are extracted from the application trace. They are evenly divided for training and testing. Figure \ref{fig:realistic}(a) shows the average SLRs across 300 test cases during the placement search. GiPH again outperforms other policies and more rapidly finds better placements within fewer search steps. Figure \ref{fig:realistic}(b) shows the SLR distribution of the final placement across test cases. GiPH is comparable to HEFT with a lower average SLR across test cases.
}

{
\paragraph{Real-world deployment and practical aspects:} 
We measure the real-world relocation cost of each task of the cooperative sensor fusion in terms of the data migration, task initialization (static data), and startup times in a small-scale deployment (Fig. \ref{fig:deployment}).  The measured relocation overhead is summarized in Table \ref{tab:deployment}.

To evaluate GiPH's practical handling of task relocation, we have integrated relocation costs into the simulator and measure the cost incurred when applying the learned policy as a certain network change occurs. The relocation cost is defined to be the sum of the data migration time and startup time. { It is worth noting that most data processing applications run recurrently to collect and process new data, similar to our sensor fusion pipeline that processes video streams and LIDAR data at a fixed frequency. Thus, a single relocation may benefit future runs of the application.}

To balance the trade-off between relocation costs and latency reduction, we divide the relocation cost by the frequency of pipeline runs. This allows us to assess the impact of relocation on pipeline performance over time by considering the potential value of task relocation in reducing the cost of multiple future runs of the application. Fig. \ref{fig:rcost}(left) illustrates the amount of relocation cost that GiPH's policy would introduce at various pipeline frequencies, based on our relocation cost model. Our analysis shows that GiPH is more likely to relocate a task at a higher cost when the pipeline runs at a higher frequency.

}

\section{Conclusion and Discussion}\label{sec:conclusion}
We present GiPH, an RL-based framework for learning generalizable placement policies that improves upon a given placement through incremental task relocation. We formulate the learning problem as a search problem such that the policy outputs incremental placement improvement steps. Using gpNet, a graph representation that captures relevant task-level and device-level features for placement, 
our framework accommodates task graphs and device networks of arbitrary size and characteristics. 
GiPH learns generalizable policies that find better placement results, faster than other search-based algorithms. 
We present a case study using realistic application traces for autonomous intersection management that utilizes the cost measurements in a real-world deployment. Although  GiPH is only  comparable to HEFT on makespan minimization, HEFT  runs slowly in practice~\citep{gcnscheduler} and is specific to makespan optimization, while GiPH can optimize a wide range of objectives defined by the reward function (e.g., resource utilization, energy cost). Fig. \ref{fig:rcost}(right) shows that GiPH outperforms random and HEFT in energy cost minimization by simply switching to a different reward function. We plan to further deploy GiPH on real-world device clusters with realistic dynamics that account for potential relocation overhead and dynmic application arrivals. Our preliminary results suggest that GiPH's results may vary depending on the stopping criterion for the placement search, and we will explore different criteria to ensure that GiPH learns a ``good'' application placement.



\section*{Acknowledgements}
This work was supported by funding from NSF grants CPS-1645578, CPS-1646235, CNS-2106891, and USDOT UTC grant 69A3551747111. We also wish to thank the reviewers for providing valuable feedback on earlier paper drafts. 

\bibliography{ref}
\bibliographystyle{mlsys2023}

%
%
%
%
%




\newpage
\appendix
\section{Artifact Appendix}

\subsection{Abstract}
The artifact contains the implementation of the learning-based placement algorithm GiPH in the paper GiPH: Generalizable Placement Learning for Adaptive Heterogeneous Computing. This artifact appendix contains information on how to use the reference implementations to reproduce the main experiment results presented in the paper. We show (1) how to create the synthetic program graph and device network datasets, (2) how to train GiPH and other baselines, and (3) how to evaluate the learned model on a test set. 

\subsection{Artifact check-list (meta-information)}
{\small
\begin{itemize}
  \item {\bf Data set:} Synthetic data, workload trace
  \item {\bf Output:} Model parameters, latency trace, intermediate sample of placement configuration
  \item {\bf Publicly available: }Yes, at \url{https://github.com/uidmice/placement-rl}
  \item {\bf Code licenses: }MIT
\item {\bf Archived: } Zenodo. DOI: 10.5281/zenodo.7879679

\end{itemize}

\subsection{Description}

\subsubsection{How delivered}
The artifact is publicly achieved using Zenodo with DOI 10.5281/zenodo.7879679 and available at \url{https://github.com/uidmice/placement-rl}.
\subsubsection{Hardware dependencies}
No hardware dependency required. All experiments in the paper are done on CPUs only, but the code allows the use of CUDA. If running on an arm mac you need to run the x86 version of conda through rosetta as DGL requires x86 architecture.
\subsubsection{Software dependencies}
This project is best run in a conda environment to satisfy version requirements, these are contained in \texttt{requirements.txt}. The prototype is done using DGL 0.9.1 with a Pytorch 1.13.0 backend.
\subsubsection{Data sets}
The method used to generate the dataset of program graphs and devices networks are detailed in Appendix \ref{sec:data}. The  implementation with detailed explanations are included  in \texttt{Generate\_data.ipynb} available in the  repository. The notebook also includes  visualization of the distribution of average communication and computation cost given the parameters. 

Upon starting the experiment, a set of parameters will be fetched from the path specified by the {\texttt{--data\_parameters}} command and used to generate the dataset for training and evaluation. The exact parameters for generating the dataset used in the paper are in  \texttt{parameters/}. By default, \texttt{parameters/single\_network.txt} is used. 

We also allow passing customized graphs and networks for training and testing. The specific datasets used for the adaptivity experiment and the  application traces from the sensor fusion pipelines are available to download: \url{https://drive.google.com/drive/folders/12nztX3XfJh3uFsTGTY2DdluQGsson2Am?usp=share_link}. 
\subsection{Installation}
Run the following commands to create a virtual environment and install relevant packages:

\begin{lstlisting}[language=bash]
  $ conda create -n placement matplotlib=3.5.1 networkx=2.5 numpy=1.19 python=3.8.10 pytorch=1.13.0 requests tqdm -c pytorch
  $ conda activate placement
  $ pip3 install simpy
  $ conda install -c dglteam dgl=0.9.1
\end{lstlisting}
\subsection{Experiment workflow}
The pipeline can be run to create a trained model and testing results by running \texttt{main.py}.

To train and evaluate the model on default parameters: 
\begin{lstlisting}[language=bash]
$ python main.py --train 
\end{lstlisting}

To train and evaluate the model with customized dataset: 
\begin{lstlisting}
$ python main.py --train 
    --load_train_graphs data/single-network/train_program.pkl 
    --load_test_graphs data/single-network/eval_program.pkl 
    --load_train_networks data/single-network/train_network.pkl 
    --load_test_networks data/single-network/eval_network.pkl
\end{lstlisting}

Each run creates a subfolder inside the log directory, \texttt{--logdir}, with a date/time stamp. The following are saved:
\begin{enumerate}
    \item The model parameters of both the GNN (\texttt{embedding\_*.pk}) and the policy network (\texttt{policy\_*.pk}).
    \item Episodic training data (\texttt{train\_data.pkl}), including the latency traces, intermediate sample placements, etc.
    \item Episodic evaluation results (\texttt{eval\_data.pkl}), if enabled, performed every after a few model updates are made.
    \item The argument inputs (\texttt{args.pkl}).
\end{enumerate}

We can also use \texttt{main.py} to run Placeto, GiPH-task-eft, and alternative GiPH implementations described in Appendix \ref{sec:ablation}, by providing additional command line arguments. 
\subsection{Evaluation and expected result}
After finish training, we can use \texttt{main.py} to load a learned model from a run directory and run a specified amount of tests on it. For evaluation, \texttt{--run\_folder} must be provided. 
\begin{lstlisting}
$ python main.py --test  
    --run_folder (logdir)/yyyy-mm-dd_hh-mm-ss_(suffix) 
    --num_testing_cases 200
\end{lstlisting}

By default, the last saved model will be loaded. To load the model parameters from an earlier save, we can do the following:
\begin{lstlisting}
$ python main.py --test  
    --run_folder (logdir)/yyyy-mm-dd_hh-mm-ss_(suffix) 
    --policy_model policy_20.pk 
    --embedding_model embedding_20.pk
\end{lstlisting}

Each test will create a subfolder inside the run directory with a name starting with \texttt{test\_} and date/time stamp. The evaluation on individual test cases (a pair of a program and a device network) will be saved individually. 

\subsection{Experiment customization}
\texttt{main.py} supports command line arguments to modify the operation and parameters of the experiment pipeline, including the learning rate, the random seed, how often we evaluate the model during training, how many times we repeat the same test cases, etc. See \texttt{README.md}.



\section{Appendix}
\begin{table*}
  \caption{Notation used throughout the paper}
  \label{tab:notation}
  \centering
  \begin{tabular*}{\textwidth}{@{\extracolsep{\fill}}ll}
    \toprule
    Symbol     & Description      \\
    \midrule
    $G$ & Task graph       \\
    $V$ & The set of nodes (task) \\
    $E$ & The set of edges (data links/dependencies) \\
    $v_i$     & Task $i$       \\
    ${\beta}_i^n,{\beta}_{ij}^e$     & Node feature of task $i$, edge feature of data link $(v_i,v_j)$, for the task graph      \\
    \midrule
    $N$ & Device network       \\
    $D$ & The set of devices \\
    $D_i$ & The set of feasible devices for task $i$ \\
    $d_k$ & Device $k$ \\
    ${b}_k^n,{b}_{kl}^e$     &  Compute feature of device $k$, communication feature between device $k$ and $l$     \\
    \midrule
    $\mathcal{M}^{G\rightarrow N}, \mathcal{M}$ & A mapping from task $V$ (of $G$) to device $D$ (of $N$)\\
    $\mathcal{P}$ & A general placement described by a tuple $(G,N,\mathcal{M}^{G\rightarrow N})$\\
    $\rho$ & The objective function as a function of the placement\\
    $S_{G,N}, A_{G,N}$ & The state space and action space of the MDP for a given problem instance\\
    $s_t,a_t,r_t$ & The state, action, and reward at step $t$\\
    $H=(V_H, E_H)$ & gpNet graph representation, with nodes $V_H$ and edges $E_H$\\
    $O_i$ & The set of placement options for a task $i$\\
    $V_{H,\mathcal{P}}$ & The set of pivots of $H$\\
    $x^n_u, x^e_{uv}$ & The composed node feature of $u$ and edge feature of $(u,v)$ in gpNet $H$\\
    $e_u$ & The embedding of node (action) $u$\\
    $q_u$ & Node (action) score of $u$\\
    \bottomrule
  \end{tabular*}
\end{table*}

\subsection{gpNet Algorithm}\label{alg:gpnet}
\begin{algorithm}[H]
\renewcommand{\thealgorithm}{}
\caption{gpNet}
\begin{algorithmic}[1]
\REQUIRE $G=(V,E,\bm{\beta}^n,\bm{\beta}^e),N=(D,\bm{b}^n,\bm{b}^e),\mathcal{M}^{G\rightarrow N}$
\STATE Initialize an empty graph $H=(V_H,E_H,\bm{x}^n,\bm{x}^e)$
\STATE $V_{H,\mathcal{P}}=\{\}$\hfill\COMMENT{The set of pivots}
\FOR {$v_i\in V$}
\STATE $O_i=\{\}$\hfill\COMMENT{The placement options of $v_i$}
\FOR {$d_j\in D_i$}
\STATE add node $u=(v_i,d_j)$ to $V_H$ and $O_i$
\STATE node feature of $u$: $x^n_u=f_n(\beta^n_i,b^n_i)$
\IF {$\mathcal{M}^{G\rightarrow N}(v_i)=d_j$}
\STATE add $u=(v_i,d_j)$ to $V_{H,\mathcal{P}}$
\ENDIF
\ENDFOR
\ENDFOR
\FOR {$(v_i,v_j)\in E$}
\FOR{$u_1=(v_i,d_k)\in O_i, u_2=(v_j,d_l)\in O_j$}
\IF{$u_1\in V_{H,\mathcal{P}}$ or $u_2\in V_{H,\mathcal{P}}$}
\STATE add edge $c=(u_1,u_2)$ to $E_H$
\STATE edge feature of $c$: $x^e_{c}=f_e(\beta^e_{ij},b^e_{kl})$
\ENDIF
\ENDFOR
\ENDFOR
\STATE \textbf{return} $H$\hfill\COMMENT{gpNet representation of the placement}
\end{algorithmic}
\end{algorithm}

\subsection{Synthetic Data Generation}\label{sec:data}
\begin{figure*}
    \centering
    \includegraphics[width=.8\linewidth]{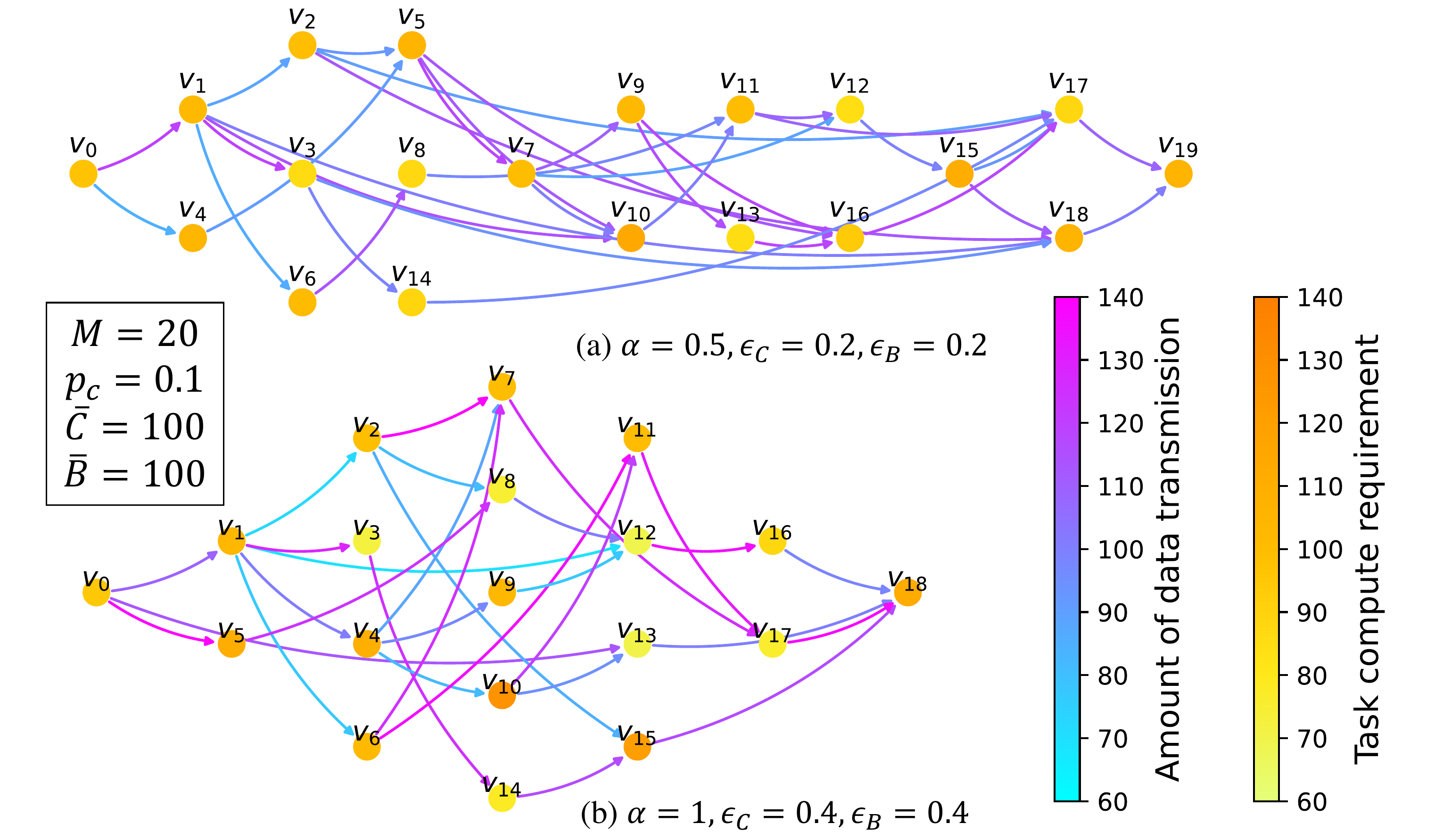}
    \caption{Example task graphs}
    \label{fig:graphs}
\end{figure*}
Following the parametric method used by Topcuoglu et al. \citep{heft}, we implement a random graph generator and a random device network generator to generate task graphs and device networks with various characteristics depending on input parameters. 

\paragraph{Task graph generator:} The task graph generator builds single-entry and single-exit task graphs, i.e., exactly one task (entry) is without any parent and exactly one task (exit) is without any child in the graph. For application graphs with more than one entry (exit) task, those tasks can be connected to a pseudo entry (exit) task with zero-cost edges. The following input parameters are used to build task DAGs:
\begin{itemize}
    \item Number of tasks in the graph $M$.
    \item Shape parameter $\alpha$. The depth of the DAG (i.e., the length of the longest path) is randomly generated from a uniform distribution with a mean value $\sqrt{M}/\alpha$. The width for each level is randomly generated from a uniform distribution with a mean value $\alpha \sqrt{M}$.
    \item Connection probability $p_c$. The probability of having a directed edge (data link) from a node at a higher level to a node at a lower level.
    \item Average compute requirement of tasks  $\overline{C}$.
    \item Average amount of data transmission along  data links  $\overline{B}$.
    \item Heterogeneity factor for compute requirements among tasks $\epsilon_{C}$. The compute requirement $C_i$ of a task $i$ is uniformly sampled from the range $\left[\overline{C}\times(1-\epsilon_{C}), \overline{C}\times(1+\epsilon_{C})\right]$.
    \item Heterogeneity factor for the amount of data transmission $\epsilon_{B}$. The amount of data to be transferred along the data link $(v_i,v_j)$, $B_{ij}$, is uniformly sampled  from the range     $\left[\overline{B}\times(1-\epsilon_{B}), \overline{B}\times(1+\epsilon_{B})\right]$.
\end{itemize}
Two example graphs with different shape parameters and heterogeneity factors are shown in Fig.  \ref{fig:graphs}. Note that a larger shape parameter ($\alpha = 1$ instead of $\alpha = 0.5$) leads to a visibly wider and shorter graph. Similarly, larger $\epsilon_B$ and $\epsilon_C$ values exhibit more variability in the amount of data transmission and task compute respectively.
\paragraph{Device network generator:} The following input parameters are used to generate device networks:
\begin{itemize}
    \item Number of devices $m$.
    \item Average compute speed of devices $\overline{SP}$.
    \item Average bandwidth between devices $\overline{BW}$.
    \item Average communication delay between devices $\overline{DL}$. The communication delay $DL_{kl}$ between device $k$ and $l$, with $k\neq l$, is uniformly sampled from $[0, 2\times \overline{DL}]$.  The delay to the device itself is zero, i.e., $DL_{kk}=0$ for all $d_k$.
    \item Heterogeneity factor for compute speed among devices $\epsilon_{SP}$. The compute speed $SP_k$ of a device $k$ is uniformly sampled from the range $\left[\overline{SP}\times(1-\epsilon_{SP}), \overline{SP}\times(1+\epsilon_{SP})\right]$.
    \item Heterogeneity factor for the communication bandwidth $\epsilon_{BW}$. The communication bandwidth $BW_{kl}$ between device $k$ and $l$, with $k\neq l$, is uniformly sampled from     $\left[\overline{BW}\times(1-\epsilon_{BW}), \overline{BW}\times(1+\epsilon_{BW})\right]$. The bandwidth for sending data to the device itself is infinite, i.e., $BW_{kk}=\infty$ for all $d_k$.
\end{itemize}

The placement constraints are specified by adding a  hardware-requirement property to each task and a corresponding hardware-support property to each device with non-zero probability.  Given a task graph and a device network, each task can only run on devices with the hardware support the task requires. A higher probability for a hardware support results in a larger number of feasible devices for a task requiring that hardware. 

 Our simulator also allows for assigning multiple values to each parameter used by the generators. A specific combination of parameter values is used to generate data. The source code of the generators are included in the supplementary material together with the parameters used for the experiments in this paper.

\subsection{Deep Learning Graphs Generation}\label{sec:dl-graphs}
To evaluate GiPH on deep learning graphs, we use ENAS~\citep{enas} to generate different neural network architectures. Specifically, we sample 10 recurrent cell designs from ENAS on the Penn Treebank benchmark for language models. One sample graph is shown in Fig. \ref{fig:rnn-cell}. For each cell design, we vary the number of unrolled steps and workload in terms of the batch size to generate 30 different deep learning graphs. The dataset contains 300 graphs in total. The number of unrolled steps is uniformly sampled from 20 to 30, and the batch size from 80 to 150. 
 \begin{figure}
     \centering
     \includegraphics[width=\linewidth]{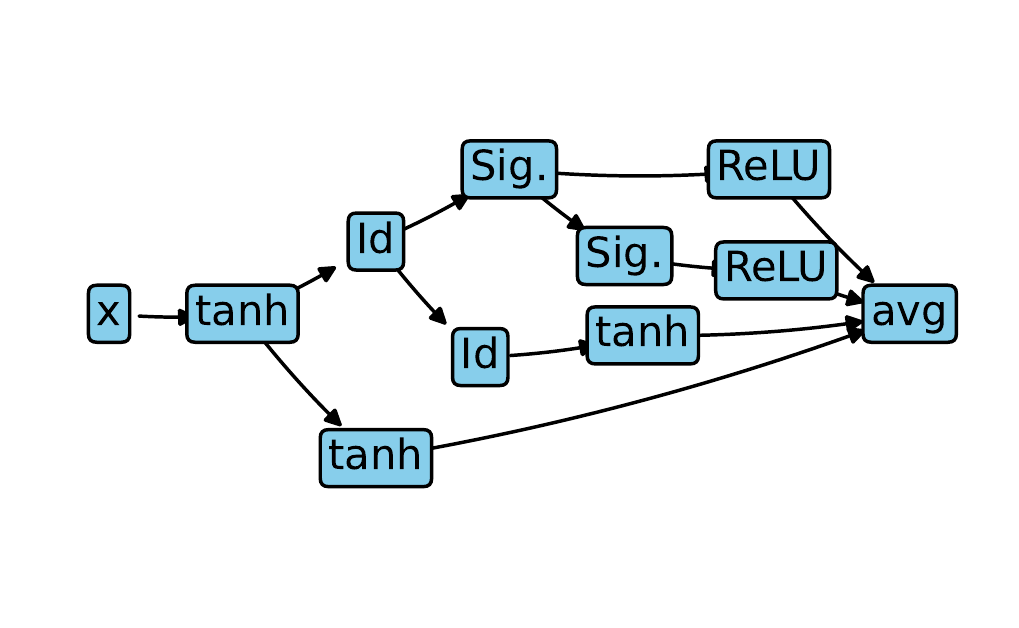}
     \caption{Sample RNN cell design from ENAS}
     \label{fig:rnn-cell}
 \end{figure}
\subsection{Case Study Details}\label{sec:case_study}

\paragraph{Task/device features:} We define an average compute requirement $\overline{C}$ for each of the four tasks  and a pair of compute features $\overline{T}$ and $\overline{S}$ for the CPU and GPU of each type of devices. $\overline{T}$ is a measure of the time used for running a unit of compute and $\overline{S}$ is a measure of the startup time of that type of device, both in milliseconds. The values are determined by fitting a computation latency model that satisfies $\overline{C_i}\overline{T_j}+ \overline{S_j}=\mu_{i,j}$, where $\mu_{i,j}$ is the measured average  time  of running task $i$ on device $j$ according to Table  \ref{tab:measurements}.

\paragraph{Communication/link features:} We estimate the sizes of the data transfer $\overline{B_{ij}}$ in bytes between tasks $i$ and $j$ based on the hardware and experiment setup as described by \citet{andert2022accurate}. We assume a communication bandwidth between devices that decays exponentially with distance $\overline{BW}=60\exp\left({-d/100}\right)$Mbps, where $d$ is the distance in meters between two devices.

\subsection{Simulator}\label{sec:simulation}
We built a Python-based runtime simulator to model a distributed computing environment with the following characteristics: (1) Each device executes runnable tasks in a first-in-first-out (FIFO) manner. (2) Task execution is non-preemptive, i.e., not interruptible by other tasks once the execution starts. (3) At most one task can run on a device at a time. (4) Computation can be overlapped with communication. The above aspects match with the real execution models in some existing distributed computing environments (e.g., Tensorflow, as demonstrated in~\citep{addanki2019placeto}). 

\paragraph{Task model:} A task is a unit of compute that can run on devices with the hardware support the task requires.  A non-entry task $v$ is \textit{runnable} on a device $d$ when all of its parents have finished execution and all of the data inputs from parents that the task takes to run are available on $d$ (i.e., the inputs to $v$ have been either locally produced on $d$ or transferred from other devices to $d$). An entry task is always runnable. The data outputs of a task $v$ become available for use on $d$ or ready to be transferred from $d$ once task $v$'s execution finishes. 

\paragraph{Device model:} Each device can execute at most one task at a time, and the task execution is assumed to be \textit{nonpreemptive}.  Each device keeps runnable tasks in a \textit{FIFO} quene and  executes them in the order they become runnable. Inter-device communication is assumed to perform without contention, and therefore, sending output data to multiple devices at the end of task execution is modeled to happen concurrently without queueing. We also assume that computation can be overlapped with communication. 

\paragraph{Latency model (synthetic data):} We model the execution (computation) time $w_{i,k}$ of running a task $v_i$ on a device $d_k$ to be proportional to the compute requirement of the task and inversely proportional to the compute speed of the device:
\begin{equation}\label{eq:w}
    w_{i,k} = \frac{C_i}{SP_k}.
\end{equation}
The data transmission (communication) time $c_{ij,kl}$ of a data link $(v_i,v_j)$, with $v_i$ mapped to device $d_k$ and $v_j$ mapped to device $d_l$, is calculated as:
\begin{equation}\label{eq:c}
    c_{ij,kl} = DL_{kl} + \frac{B_{ij}}{BW_{kl}}.
\end{equation}
Note that the communication time is zero if the two tasks are placed on the same device because we set  zero delay and infinite bandwidth for local data transmission when generating the device network (Appendix~\ref{sec:data}). 

With non-zero noise $\sigma\in(0,1)$, Equations \ref{eq:w} and \ref{eq:c} give the expected computation and communication times, but their  realizations are sampled from a uniform distribution with range $[w_{i,k}(1-\sigma), w_{i,k}(1+\sigma)]$ for computation time and range $[c_{ij,kl}(1-\sigma), c_{ij,kl}(1+\sigma)]$ for communication time. In a more realistic setting, the latency model can also be replaced by real measurements of communication and computation times from profiling tools.

A discrete event simulation tool for Python, SimPy, is used to handle the interactions between the start and end of data transmission/task execution across devices.  
Each device $d_k$ keeps a FIFO  queue $Q_k$ for all runnable tasks on it. The  queue is dequeued when $d_k$ is not busy or have finished executing the previous task.
We define the following events with corresponding event handlers:
\begin{itemize}
    \item \textit{Task start} $t_i^s$: Used to mark the start time of the execution of a task $v_i$. Given $\mathcal{M}(v_i)=d_k$, its timestamp $t_i^s$ is the time when the task is dequeued from $Q_k$. Device $d_k$ becomes busy at $t_i^s$ and stays busy until $t_i^s+w_{i,k}$.
    \item \textit{Task done} $t_i^d$: Used to mark the end of the execution of a task $v_i$. Given $\mathcal{M}(v_i)=d_k$, its timestamp $t_i^d=t_i^s+w_{i,k}$. The device becomes free (not occupied) when the task is done.  This event also triggers the data transmission to other devices where the child tasks reside, i.e., data communication from $d_k$ to  $\mathcal{M}(v_j)$ for $(v_i,v_j)\in E$.
    \item \textit{Transmission start} $t_{ij}^s$: Used to mark the start of the transmission of the data link $(v_i,v_j)$. Its timestamp is the same as the \textit{task done} event of $v_i$, i.e., $t_{ij}^s=t_i^d$ for all $(v_i,v_j)\in E$.
    \item \textit{Transmission done} $t_{ij}^d$: Used to mark the end of the transmission of the data link $(v_i,v_j)$. Given $\mathcal{M}(v_i)=d_k$ and $\mathcal{M}(v_j)=d_l$, the timestamp  $t_{ij}^d=t_{ij}^s + c_{ij,kl}$.
    \item \textit{Task runnable} $t_i^r$: Used to signal  a task $v_i$ becomes runnable after receiving all inputs, i.e., $t_i^r=\max_{(v_j,v_i)\in E}t_{ji}^d $. When a task $v_i$ becomes runnable, it is inserted to the FIFO queue of the device $d_k$ where the task is placed (i.e., $\mathcal{M}(v_i)=d_k$).
\end{itemize}
After the entry task starts and the data flows through the whole graph, the completion time (makespan) of the task graph is the time duration from the start of the entry task to the end of the exit task, i.e.,  $t^d_{exit}-t^s_{entry}$.

\subsection{Implementation Alternatives}\label{sec:ablation}
\begin{figure*}[h]
    \centering
    \includegraphics[width=\linewidth]{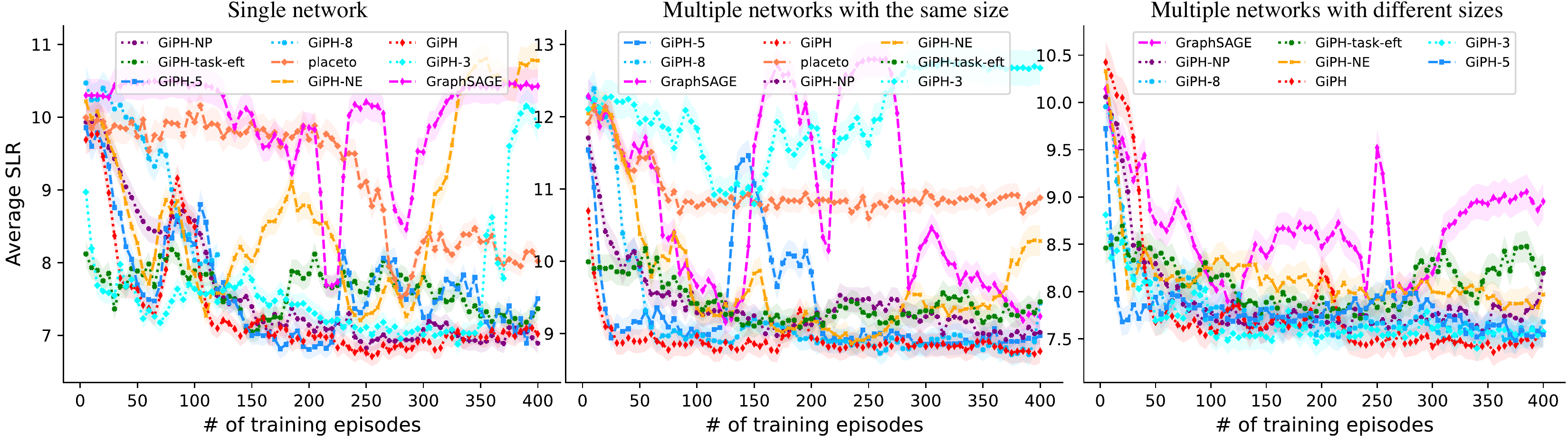}
    \caption{Average SLR across 20 evaluation cases with respect to the number of training episodes.  Left: Training and evaluating on a single network. Middle: Training and evaluating on fixed-sized device networks. Right: Training and evaluating on device networks of various sizes. }
    \label{fig:convergence}
\end{figure*}

 We compare with the following GNN alternatives:
\begin{enumerate}[label=\textbf{\arabic*}.]
    \item \textbf{GiPH-NE}: GiPH's two-way message passing without edge features. To compensate for the loss of edge information, the mean feature value of out edges of a node is appended to its node feature.
\item \textbf{GraphSAGE-NE}: With the same node features as used by GiPH-NE, GraphSAGE-NE replaces the two-way message passing with a 3-layer uni-directional GraphSAGE network~\citep{graphsage}.
\item \textbf{GiPH-NE-Pol}: GiPH without GNN. GiPH-NE-Pol directly feeds the same raw node features as used by GiPH-NE to the policy network. 
\end{enumerate}
While combining per-edge information with node features, these implementations still use the gpNet as inputs.  The effect of using gpNet can be seen by comparing to  GiPH-task-eft, which does not use gpNet to encode the placement information.
\begin{enumerate}[label=\textbf{\arabic*}.]
\setcounter{enumi}{3}
    \item \textbf{GiPH-task-eft}: GiPH task selection with EFT device selection. Without using gpNet, selecting a task and deciding where to place it  are done separately. 
\end{enumerate}

The number of message passing steps in GiPH is equal to the depth of the graph, which can be time-consuming for large graphs. We consider the following GNN alternative that limits the number of steps to propagate local structural information.
\begin{enumerate}[label=\textbf{\arabic*}.]
\setcounter{enumi}{4}
    \item \textbf{GiPH}-\textit{k}: GiPH $k$-step two-way message passing defined by:
    \begin{align}
        e^{t+1}_u=h^c_2\left(\sum_{v\in \xi(u)}h^c_1\left([e^t_v\mathbin\Vert x_{vu}^e]\right)\right) + h_3^c(x_u^n),
    \end{align}
 for $t=0,...,k$ with $e^{0}_u = x_u^n$. $h^c_1(\cdot)$, $h^c_2(\cdot)$ and $h^c_3(\cdot)$ are feed-forward neural networks with trainable parameters shared for message passing steps in each direction. 
\end{enumerate}

\paragraph{Neural network implementation:} 

For the k-step message passing GiPH-$k$, a two-layer FNN ($h_3^c$) with the same number of hidden units as the input dimension is used to pre-embed node features to a higher-dimensional space. We set the output dimension to be $10$. Similar to GiPH, the message passing and message aggregation use single-layer FNNs. Details are included in Table \ref{tab:nn_dim} and Table \ref{tab:nn_details}.

\begin{table*}
    \caption{Dimension Summary}
     \label{tab:nn_dim}
    \centering
    \vskip 0.1in
    \begin{tabular}{cccccccc}
    \toprule
         &  GiPH & GiPH-k & \makecell{GiPH-\\ NE} & \makecell{GiPH-\\NE-Pol} & \makecell{GraphSAGE- \\ NE} & Placeto & \makecell{RNN-\\ Placer}\\\cmidrule(r){2-8}
         \makecell{Node \\ feature\\$dim_n$} & \makecell{4} & 4 & 8 & 8 & 8 & 5 & -\\\midrule
         \makecell{Edge \\ feature\\$dim_e$} & \makecell{4} & 4 & - & - & - & - & -\\\midrule
         \makecell{Embed- \\ ding  \\ $dim_o$} & \makecell[b]{$5$} & \makecell{5} & \makecell{5} & \makecell{ - } & \makecell{ 10 } & \makecell{5} & \makecell{$n_{type}+1$ \\ $+\max(d_{out})$ \\$+n_{nodes}$}\\\bottomrule
    \end{tabular}
    \vskip -0.1in
\end{table*}

\begin{table*}
  \caption{Neural network implementation details.}
  \label{tab:nn_details}
  \centering
  \begin{tabular}{ccccccc}
    \toprule
     & \makecell{Node \\ transform\\ layer} & \makecell{Message \\ function} & \makecell{Aggregation\\ function} & \makecell{Message\\ passing \\$k$} & \makecell{Node\\ summery\\ dim} & Policy \\   \cmidrule(r){2-7}
     \makecell{ GiPH}  &
    \makecell{$\mathbf{dim_n}$ \\ $dim_n$\\ $\mathbf{dim_o}$  } & 
    \makecell{$\mathbf{dim_o+dim_e}$ \\ $\mathbf{dim_o+dim_e}$  }&
    \makecell{$\mathbf{dim_o+dim_e}$ \\ $\mathbf{dim_o}$  }&
    \makecell{Graph\\depth} &
    \makecell{${dim_o*2}$ \\$=10$} &
    \makecell{$\mathbf{10}$ \\ $16$\\ $\mathbf{1}$  } \\\midrule
      \makecell{ GiPH-k}  &
    \makecell{$\mathbf{dim_n}$ \\ $dim_n$\\ $\mathbf{dim_o}$  } & 
    \makecell{$\mathbf{dim_o+dim_e}$ \\ $\mathbf{dim_o+dim_e}$  }&
    \makecell{$\mathbf{dim_o+dim_e}$ \\ $\mathbf{dim_o}$  }&
    \makecell{$k$} &
    \makecell{${dim_o*2}$ \\$=10$} &
    \makecell{$\mathbf{10}$ \\ $16$\\ $\mathbf{1}$  } \\\midrule
     \makecell{ GiPH-\\ NE}  &
    \makecell{- } & 
    \makecell{$\mathbf{dim_n}$ \\ $\mathbf{dim_n}$  }&
    \makecell{$\mathbf{dim_n}$ \\ $\mathbf{dim_o}$  }&
    \makecell{Graph\\depth} &
    \makecell{${dim_o*2}$ \\$=10$} &
    \makecell{$\mathbf{10}$ \\ $16$\\ $\mathbf{1}$  } \\\midrule
     \makecell{ GiPH-\\ NE-Pol}  &
    \makecell{- } & 
    \makecell{- }&
    \makecell{- }&
    \makecell{-} &
    \makecell{-} &
    \makecell{$\mathbf{8}$ \\ $16$\\ $\mathbf{1}$  } \\\midrule
     \makecell{ Graph- \\ SAGE\\-NE}  &
    \makecell{$\mathbf{dim_n}$ \\  $\mathbf{16}$ } & 
    \makecell{$\mathbf{16}$ \\ $\mathbf{16}$  }&
    \makecell{$\mathbf{16}$ \\ $\mathbf{dim_o}$  }&
    \makecell{$3$} &
    \makecell{${dim_o}$ \\$=10$} &
    \makecell{$\mathbf{10}$ \\ $16$\\ $\mathbf{1}$  } \\\midrule
     \makecell{ Placeto}  &
    \makecell{$\mathbf{dim_n}$ \\ $dim_n$\\ $\mathbf{dim_n}$  } & 
    \makecell{$\mathbf{dim_n}$ \\ $dim_n$\\ $\mathbf{dim_n}$  }&
    \makecell{$\mathbf{dim_n}$ \\ $dim_n$\\ $\mathbf{dim_n}$ }&
    \makecell{$8$} &
    \makecell{${dim_o}*2*4$ \\$=40$} &
    \makecell{$\mathbf{40}$ \\ 32\\ $\mathbf{n_{dev}}$  }  \\
    \bottomrule
  \end{tabular}
\end{table*}
\paragraph{Policy convergence:} We train the policies using different implementations with the same training dataset for $200$ training episodes  and test the policy convergence by evaluating the learned policies  every $5$ training episodes. The evaluation is done by applying the learned policies to the same set of $20$ evaluation cases, whose task graphs and device networks are not in the training dataset.

The experiment results are shown in Fig. \ref{fig:convergence}. For the result on the right hand side, we use the same device network throughout the training and testing. For the middle plot, we use fixed-sized device networks, and for plot on the right hand side, we further vary the size of the device networks. For the first two experiments, we also include Placeto for comparison.  We find that the policies tend to vary less when trained on device networks of various sizes, which suggests the benefit of having diverse training data. 

The policies of GiPH, GiPH-3, GiPH-5 and GiPH-NE-Pol converge in both cases. GiPH-task-eft fails to converge in both cases, probably because it does not have a unified placement update policy without using gpNet. The separation of the device selection with the RL policy for task selection makes the policy learning harder.  GraphSAGE-NE and GiPH-NE both incorporate edge features into the node features; while GiPH-NE constructs  message passing in both forward and backward directions, the message passing of GraphSAGE-NE is uni-directional, which may be the cause of divergence of GraphSAGE-NE in both cases.

\begin{figure}[h]
    \centering
    \includegraphics[width=\linewidth]{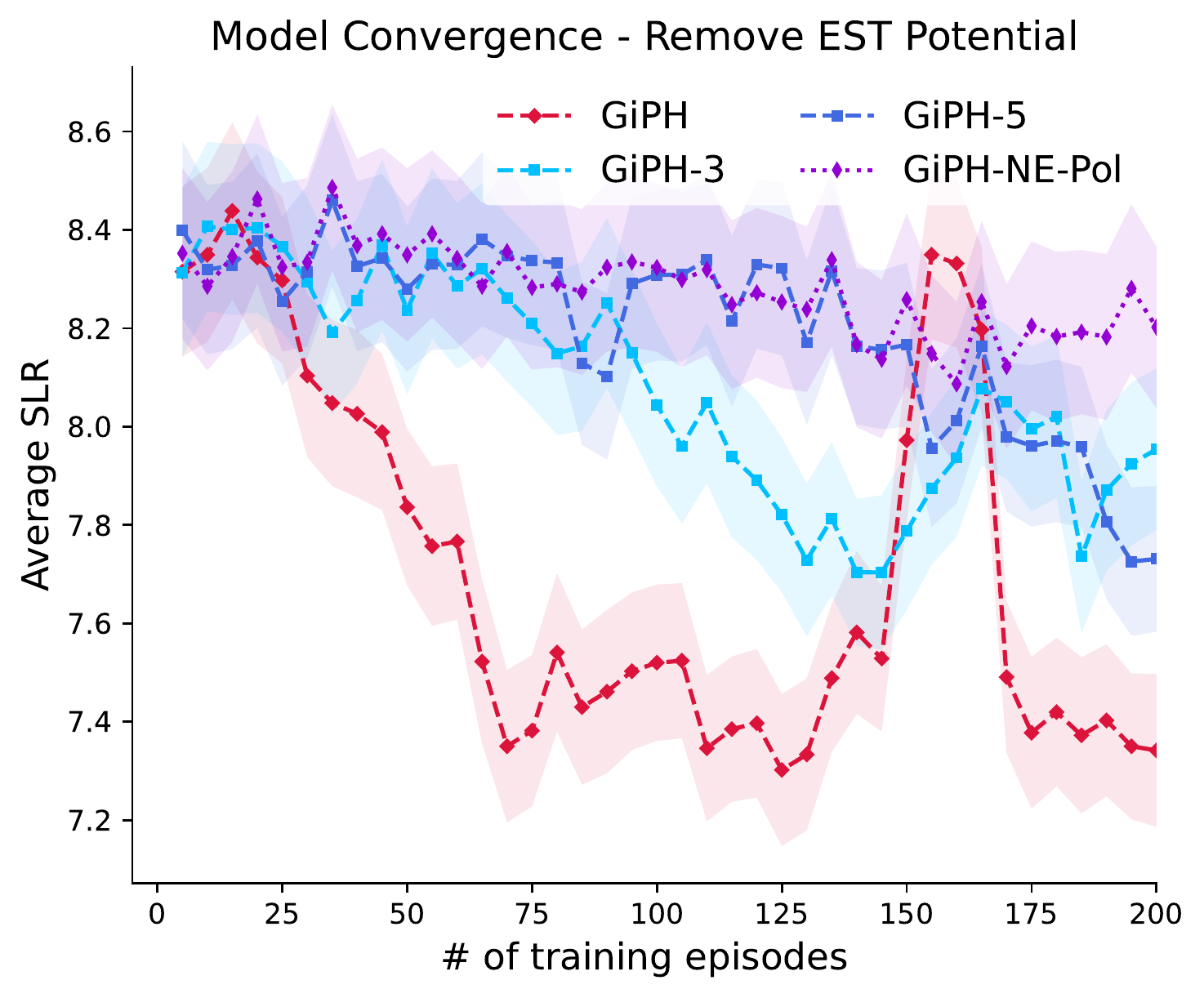}
    \caption{Convergence result after removing the start-time potential as a node feature.}
    \label{fig:conv_n}
\end{figure}
We attribute the success of GiPH-NE-Pol to our feature selection described in Appendix \ref{sec:training_detail}. Specifically, the start-time potential as a node feature itself provides aggregated information of the neighbors, which may greatly help the policies to converge. We further repeat the multisized-device-network  experiment without using the start-time potential as a node feature. The convergence of GiPH, GiPH-3, GiPH-5 and GiPH-NE-Pol is shown in Fig. \ref{fig:conv_n}. In this case,  without using GNN, GiPH-NE-Pol performs poorly and the policy does not improve the average SLR at all. In contrast, GiPH still successfully creates a sharp drop of SLR at the beginning of the training. The training efficiency of all four policies decreases after removing the start-time potential node feature, with GiPH being the least-influenced.

\paragraph{Placement quality:} We randomly select 1000 test cases  from the test dataset to test learned policies. We count the number of occurrences of better placements among GiPH, its variants and HEFT. The result is summarized in Table \ref{tab:plcement-compare}. In general, GiPH is better than its variants and produces comparable results as HEFT. 
\begin{table*}
  \caption{Pair-wise placement quality comparison, showing the percentage of test cases for which the method in the row index has SLR that is better than/equal to/worse than the SLR for the method in the column index. For example, GiPH's SLR is better than GiPH-3's for 53.0\% of the test cases and equal to GiPH-3's for 7.0\% of the test cases.}
  \label{tab:plcement-compare}
  \centering
  \begin{tabular}{llP{1.2cm}P{1.2cm}P{1.2cm}P{1.2cm}P{1.2cm}P{1.2cm}P{1.2cm}}
    \toprule
     & & GiPH&   GiPH-3 & GiPH-5 & GiPH-NE & GiPH-NE-Pol & GiPH-task-eft & HEFT  \\   \cmidrule(r){3-9}
    \multirow{3}{1.1cm}{GiPH}
    & Better & & $\textbf{53.0\%}$ & {$\textbf{55.2\%}$} & {$\textbf{74.8\%}$} & {$\textbf{60.6\%}$} & {$\textbf{82.2\%}$} & {$\textbf{59.0\%}$} \\
    & Equal & & $7.0\%$     & $6.4\%$   & $2.4\%$   & $6.0\%$   & $3.0\%$   & $5.2\%$ \\
    & Worse & & $40.0\%$    & $38.4\%$  & $22.8\%$  & $33.4\%$  & $14.8\%$  & $35.8\%$ \\\midrule
    \multirow{3}{1.1cm}{GiPH-3}
    & Better & $40.0\%$  &          & $\textbf{51.6\%}$ & $\textbf{68.8\%}$ & $\textbf{55.8\%}$ & $\textbf{74.6\%}$ & $\textbf{51.4\%}$ \\
    & Equal  & $7.0\%$           &  & $5.0\%$   & $1.8\%$   & $5.3\%$   & $3.0\%$   & $4.4\%$ \\
    & Worse  & $\textbf{53.0\%}$ &  & $43.4\%$  & $29.4\%$  & $38.9\%$  & $22.4\%$  & $44.2\%$ \\\midrule
    \multirow{3}{1.1cm}{GiPH-5}
    & Better&  $38.4\%$     &  $43.4\%$     &           & $\textbf{66.0\%}$ & $\textbf{52.0\%}$ & $\textbf{74.8\%}$ & $\textbf{51.2\%}$ \\
    & Equal & $6.4\%$       & $5.0\%$       &               &$1.8\%$    & $3.6\%$ & $2.2\%$ & $3.0\%$ \\
    & Worse & {$\textbf{55.2\%}$} & $\textbf{51.6\%}$  &    & $32.2\%$  & $44.4\%$ & $23.0\%$ & $45.8\%$ \\\midrule
\multirow{3}{1.1cm}{GiPH-NE}
    & Better  &  $38.4\%$    & $29.4\%$ &  $32.2\%$ &   & ${32.6\%}$ & $\textbf{61.1\%}$ & ${32.6\%}$ \\
    & Equal & $2.4\%$       & $1.8\%$ & $1.8\%$&  & $1.4\%$ & $0.7\%$ & $2.2\%$ \\
    & Worse & {$\textbf{74.8\%}$}  & $\textbf{68.8\%}$  &$\textbf{66.0\%}$ &  & $\textbf{66.0\%}$ & $38.2\%$ & $\textbf{65.2\%}$ \\\midrule
\multirow{3}{1.1cm}{GiPH-NE-Pol}
    & Better&$33.4\%$   &  $38.9\%$     &  $44.4\%$   &  $\textbf{66.0\%}$ &  & $\textbf{70.4\%}$ & ${44.6\%}$ \\
    & Equal &$6.0\%$   & $5.3\%$        & $3.6\%$  & $1.4\%$ & & $2.6\%$ & $2.8\%$ \\
    & Worse &{$\textbf{60.6\%}$}  & $\textbf{55.8\%}$  & $\textbf{52.0\%}$ &  ${32.6\%}$  & & $27.0\%$ & $\textbf{52.6\%}$ \\\midrule
    \multirow{3}{1.1cm}{GiPH-task-eft}
    & Better&$14.8\%$  & $22.4\%$ & $23.0\%$ &$38.2\%$ & $27.0\%$ &  & ${29.2\%}$ \\
    & Equal & $3.0\%$   &$3.0\%$ &  $2.2\%$ & $0.7\%$& $2.6\%$ &  & $6.4\%$ \\
    & Worse &{$\textbf{82.2\%}$}  &$\textbf{74.6\%}$  &$\textbf{74.8\%}$ & $\textbf{61.1\%}$  &$\textbf{70.4\%}$ &  & $\textbf{64.4\%}$ \\
    \bottomrule

  \end{tabular}
\end{table*}

\subsection{Experiment Details}\label{sec:training_detail}
\paragraph{Features:} The node feature vector of $(v_i,d_k)$ in gpNet consists of: (1) the compute requirement of the task $C_i$, (2) the compute speed of the device $SP_k$, (3) the expected compute time $w_{i,k}$, and (4) the start-time potential of task $v_i$ on $d_k$, which is defined as the time difference between the earliest possible start time of $v_i$ on $d_k$ and the actual start time of $v_i$. 

The edge feature vector of $((v_i,d_k),(v_j,d_l))$ in gpNet consists of: (1) the amount of data transmission  from $v_i$ to $v_j$,  $B_{ij}$, (2) the communication bandwidth from device $d_k$ to $d_l$, $BW_{kl}$, (3) the communication start-up delay $DL_{kl}$, and (4) the expected communication time $c_{ij,kl}$.

For Placeto,  the node feature vector of each operator is created by concatenating (1) the average compute time, (2) the average output data bytes, (3) the current placement, (4) an indicator of whether the operator is the current one to be placed, and (5) an indicator of whether the operator has been placed in the episode. 

For the RNN-based placer, the input embedding of each operator is created by concatenating four vectors: (1) a one-hot encoding of the type of hardware requirement for placement constraints, (2) a scalar of its compute requirement, (3) a vector containing the number of data bytes of all its outgoing edges, of size equal to the maximum out-degree of the graph, and (4) a vector for the adjacency of the operator, of size equal to the number of operators in the graph. The dimensions of the node feature, edge feature, and embedding are summarized in Table \ref{tab:nn_dim}.

\paragraph{RL training:}\label{sec:train}
The  policy gradient method REINFORCE is used  for training the RL policy~\citep{reinforce}. During each episode, a placement problem $(G, N)$ is sampled from a training set $\mathcal{G}_T\times \mathcal{N}_T$. Starting from a random placement $s_0$, the agent collects observations $(s_t,a_t,r_t)$ at each step $t=0,...,T$ following the current policy $\pi_\theta$. It updates its policy parameters at the end of each episode
\begin{equation*}\label{eq:reinforce}
    \theta\leftarrow \theta + \alpha \sum_{t=0}^{T}\gamma^t\nabla_\theta\log \pi_\theta (a_t|s_t)\left(\sum_{t'=t}^{T}\gamma^{t'-t}r_{t'}-b_{t}\right),
\end{equation*}
where $\alpha$ is the learning rate, $\gamma$ is the discounting factor, and $b_t$ is a baseline for reducing the variance of the policy gradient~\citep{rl_baseline}. 
{$b_t$} can be any function as long as it does not depend on the action {at time $t$}. We set it to be the average reward before step $t$ in an episode. 

\paragraph{Running time:} We report the average training and running times of each policy. All experiments are done on CPUs only. Both the training time and the running time include the time used to generate the input graphs (with features) and run the policy (for placement updates), and training has additional gradient update steps at the end of each episode. The values reported in Table \ref{tab:time} are averaged over placement samples taken during the training and testing (given the same training and testing datasets). Fig. \ref{fig:time} shows how the policy running and training times vary with the size of the application graphs. Since in GiPH the message passing runs sequentially from entry node to the exit node, the policy running time of GiPH grows with the size of the input graph. Limiting the number of message passing steps to $k$ (GiPH-3, GiPH-5) significantly reduces the  overhead of running the policy. 
\begin{table*}
    \centering
    \caption{Policy Running Time per Placement Sample}
     \label{tab:time}
    \begin{tabular}{P{2cm}P{1.1cm}P{1.1cm}P{1.1cm}P{1.3cm}P{1.3cm}P{1.3cm}P{1.3cm}} 
    \toprule
         &  GiPH & GiPH-3 & GiPH-5 & GiPH-NE & GiPH-NE-Pol & Graph-SAGE-NE & Placeto\\\cmidrule(r){2-8}
         Training time per placement sample (sec) & $0.565\pm 0.353$ & $0.145\pm 0.049$ & $0.178\pm 0.064$ & $0.360\pm 0.257$ & $0.027\pm 0.011$ & $0.157\pm 0.075$ & $0.255\pm 0.040$\\\midrule
         Running time per placement sample (sec)  & $0.340\pm 0.256$ & $0.114\pm 0.026$ & $0.132\pm 0.035$ & $0.240\pm 0.187$ & $0.027\pm 0.008$ & $0.138\pm 0.051$ & $0.162\pm 0.051$\\\bottomrule
    \end{tabular}
\end{table*}

\begin{figure*}
    \centering
    \includegraphics[width=\linewidth]{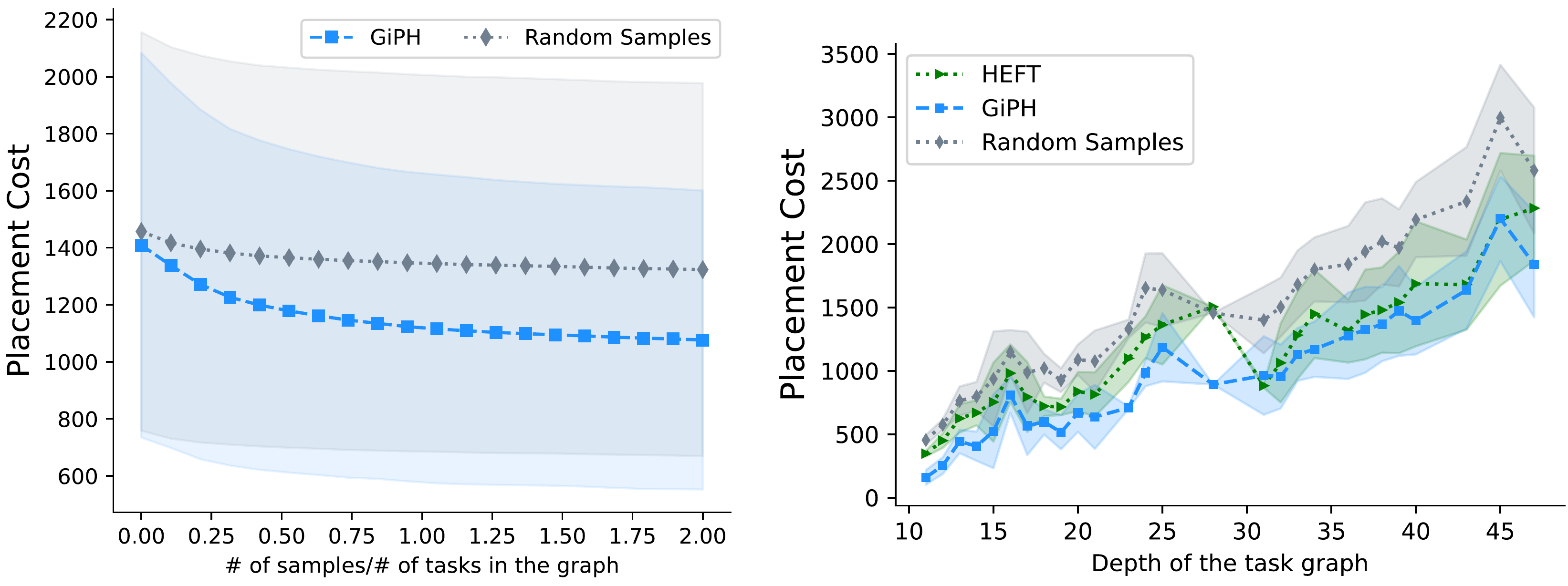}
    \caption{The total communication and computation cost of the placement found by HEFT, GiPH, and random sampling, as a function of depth of the task graph. }
    \label{fig:cost_minimization}
\end{figure*}

\begin{figure*}
    \centering
    \includegraphics[width=\linewidth]{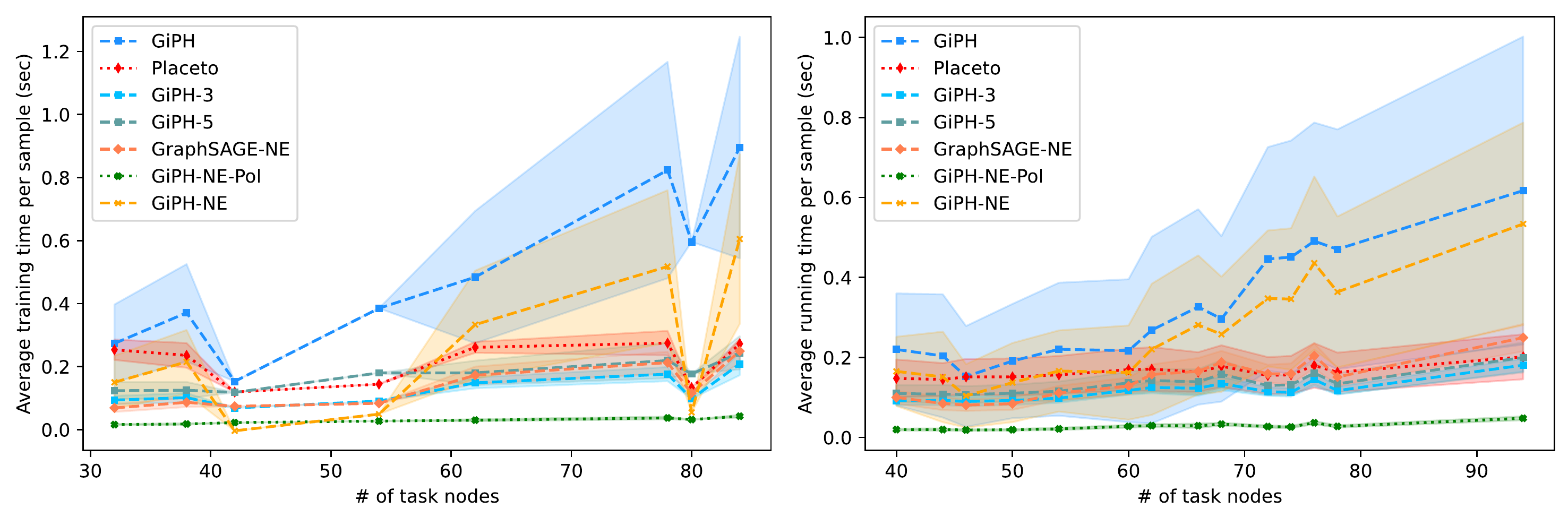}
    \caption{Left: Average training time averaged over all placement samples taken during  training. Right: Average policy running time averaged over all placement samples taken during  testing. }
    \label{fig:time}
\end{figure*}

\subsection{Supplementary Results}\label{sec:supplemental}


\paragraph{Total cost minimization:}  To demonstrate that GiPH provides a general learning framework for optimizing a variety of objectives (e.g., resource utilization and balancing, energy cost, etc.), we also test GiPH performance for cost minimization, where the cost is defined as the sum of communication cost of each data link and computation cost of each task, i.e., $Cost=\sum_{i\in V} w_{i,\mathcal{M}(i)} + \sum_{ij\in E} c_{ij,\mathcal{M}(i)\mathcal{M}(j)}$ minimized over feasible mapping $\mathcal{M}$. 

We use the same training and testing datasets as in the multiple-device-network case, and simply replace the reward with the cost reduction at each step. Fig. \ref{fig:cost_minimization} (left) shows the search efficiency for cost minimization compared with a random sampling baseline and Fig. \ref{fig:cost_minimization} (right) reports the total cost of the final placements across testing cases found by GiPH, random sampling, and HEFT.





\end{document}